\documentclass{article} 
\usepackage{iclr2027_conference,times}

\usepackage{amsmath,amsfonts,bm}

\def\eqref#1{equation~\ref{#1}}

\def\1{\bm{1}}

\DeclareMathAlphabet{\mathsfit}{\encodingdefault}{\sfdefault}{m}{sl}
\SetMathAlphabet{\mathsfit}{bold}{\encodingdefault}{\sfdefault}{bx}{n}

\usepackage{hyperref}
\usepackage{url}
\hypersetup{
  colorlinks=true,
  urlcolor=blue 
}

\usepackage{amssymb}
\usepackage{booktabs}
\usepackage{multirow}
\usepackage{amsmath}
\usepackage{amssymb}
\usepackage{amsthm}
\usepackage[table]{xcolor}
\usepackage{caption}

\usepackage{tikz}
\usetikzlibrary{positioning, arrows.meta}

\title{Joint Subspace Sparse Attention for Efficient Long-Context Decoding}

\author{Jiwon Song\thanks{Equal Contribution} ,  Yoongon Kim\footnotemark[1] ,  Jae-Joon Kim \\
Seoul National University\\
\texttt{\{jiwon.song,yoon\_g\_kim,kimjaejoon\}@snu.ac.kr}\\
\url{https://github.com/jiwonsong-dev/JSSA}
}

\iclrfinalcopy 
\begin{document}

\maketitle

\begin{abstract}
Sparse attention reduces the KV-cache traffic that dominates long-context LLM decoding, but practical methods largely operate at coarse granularity.
Token-level selection offers finer control, yet presents two challenges: low-dimensional scoring may lose query--key structure needed for accurate ranking, while gathering scattered selected tokens can offset the resulting latency benefit.
We introduce Joint Subspace Sparse Attention (JSSA) to address both challenges.
For selection quality, JSSA learns a shared low-dimensional query--key subspace by minimizing token-score preservation error.
We upper-bound the resulting non-convex objective with a tractable marginal objective that admits a closed-form spectral solution; direct refinement of the exact objective provides no measurable improvement in the tested settings.
For execution efficiency, we develop a gather-free token-level indexed-attention kernel that supports distinct per-KV-head selections under the standard grouped-query attention (GQA) KV-cache layout.
Across standard GQA, hybrid sliding-window/global attention, and Multi-head Latent Attention (MLA) models, JSSA consistently remains close to full-KV accuracy while avoiding the substantial generation-length inflation and architecture-dependent failures observed with competing methods.
On an NVIDIA B200 GPU, JSSA achieves up to $6.43\times$ the decoding throughput of full-KV attention at 128K context.
JSSA achieves reasoning accuracy comparable to model-native DSA without model training and reaches 96--99\% of its decoding throughput even at 700B scale large models.
\end{abstract}

\section{Introduction}
\label{sec:intro}

Long-context autoregressive decoding is often bottlenecked by repeatedly accessing the growing KV cache.
With full attention, each decoding step scores the current query against the entire context and reads all the corresponding values, causing memory traffic to grow linearly with sequence length.
Because attention is typically memory-bandwidth-bound in this regime, reducing KV accesses is one of the most direct ways to accelerate long-context decoding.
Sparse attention reduces KV access by selecting only part of the context at each decoding step.
Block- or page-level methods~\citep{quest,seerattentionr} align naturally with paged KV-cache layouts and block-sparse kernels, whereas token-level methods~\citep{tokenselect,sparq} such as Loki~\citep{loki} and FASA~\citep{fasa} select individual positions.
Token-level selection can represent finer-grained sparse patterns, yet this flexibility has not translated reliably into either higher accuracy or lower latency.
\begin{figure}[t]
\centering
\vspace{-16pt}
\includegraphics[width=0.95\textwidth]{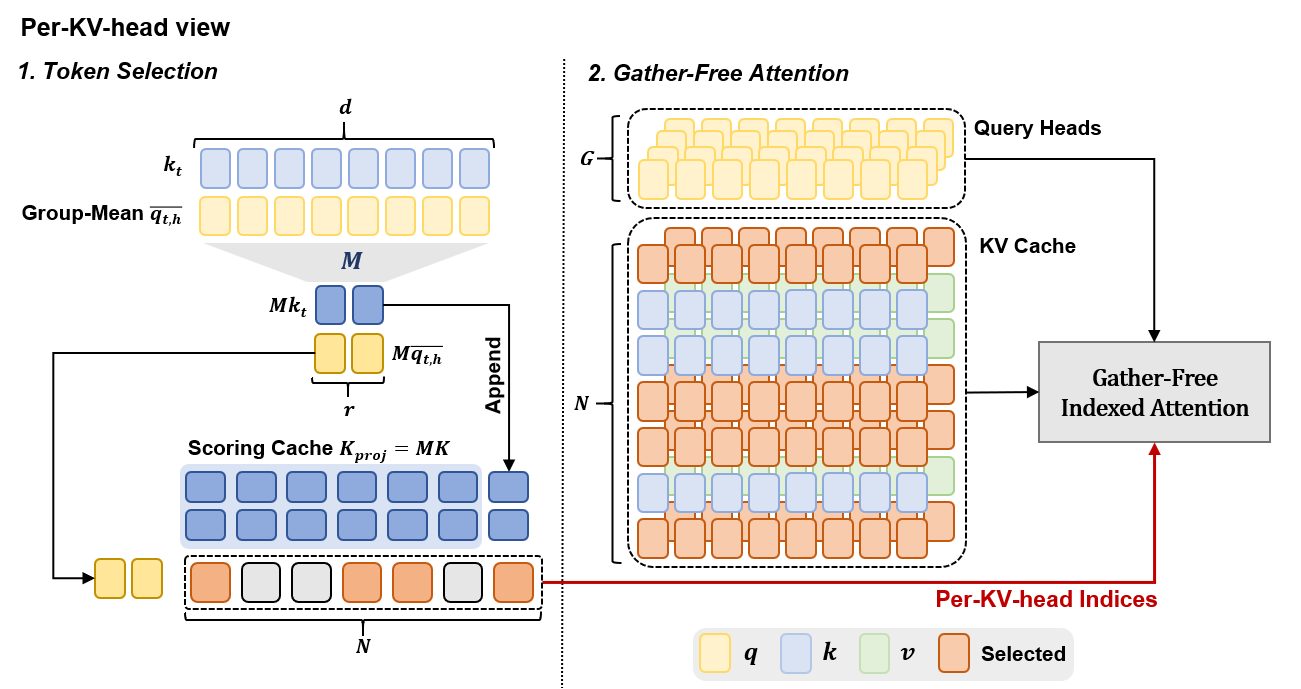}
\vspace{-8pt}
\caption{
\textbf{JSSA at one decoding step for a single KV head.}
The mean of its associated query heads scores a projected-key cache in a shared low-dimensional subspace to select the top-$B$ token positions.
The query heads share this index set, and gather-free indexed attention accesses the corresponding original full-dimensional keys and values.
}
\vspace{-8pt}
\label{fig:overview}
\end{figure}
This finer granularity creates two challenges.
First, objectives that preserve queries or keys independently may not preserve the bilinear scores used for token ranking.
Second, token-level selection produces irregular per-KV-head index sets in grouped-query attention (GQA).
Gathering them into a contiguous buffer adds overhead, and forcing all KV heads to share one selected set can degrade accuracy.

We introduce Joint Subspace Sparse Attention (JSSA), a training-free method that addresses both challenges.
For selection, JSSA learns a shared query--key subspace by minimizing bilinear score error through a closed-form spectral surrogate; directly refining the exact objective provides no measurable selection benefit in the tested settings.
For execution, JSSA preserves distinct per-KV-head token sets and consumes them directly through a gather-free indexed-attention kernel under the standard GQA KV-cache layout.
Together, these components provide an end-to-end token-level sparse-decoding path across GQA and MLA~\citep{deepseekv2mla} models.

Our main contributions are:
\begin{itemize}
    \item \textbf{Score-preserving joint query--key projection.}
    We formulate an exact bilinear score-error objective, derive a tractable marginal upper bound with a closed-form spectral solution, and show that direct refinement provides no measurable selection benefit in the tested settings.

    \item \textbf{Gather-free per-KV-head token-level attention for GQA.}
    We design, to the best of our knowledge, the first indexed-attention kernel that supports distinct per-KV-head selections under the standard GQA KV-cache layout without physically gathering selected entries.

    \item \textbf{Validation across architectures and scales.}
    JSSA remains close to full-KV accuracy across GQA, hybrid local/global attention, and MLA models, and achieves up to $6.43\times$ the decoding throughput of full-KV attention at 128K context on a B200 GPU. 
    We further show that JSSA attains reasoning accuracy comparable to model-native DSA~\citep{deepseekv32} on 700B-parameter scale models, without any model training.
\end{itemize}

\section{Motivation}
\label{sec:motivation}

\subsection{Low-Dimensional Token Scoring Requires the Right Objective}
\label{sec:motivation_projection}

Token-level selection must score every candidate token.
Unlike page-level methods, which reduce cost along the sequence axis by assigning one score to each page, token-level methods retain every position as a candidate and must instead reduce the channel dimension used for scoring.
They may exploit part of original channels, as in FASA, or project the scoring vectors onto an arbitrary low-rank subspace, as in Loki.

\begin{figure}[t]
\centering
\vspace{-16pt}
\includegraphics[width=1.0\textwidth]{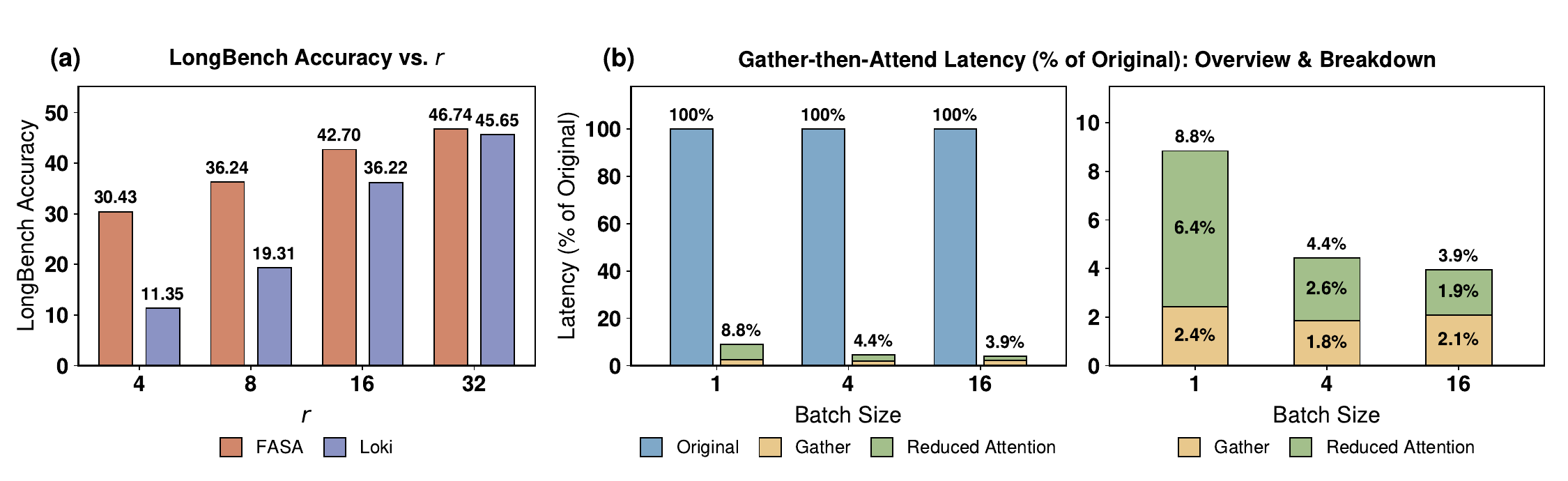}
\vspace{-22pt}
\caption{
\textbf{Low-dimensional scoring and gather overhead on LLaMA-3.1-8B-Instruct.}
(a) LongBench accuracy versus channel budget $r$ for FASA and Loki.
(b) Gather-then-attend latency at 128K context, normalized to dense attention, with a breakdown of the residual sparse-attention cost.
}
\vspace{-8pt}
\label{fig:motivation}
\end{figure}

Discrete channel selection is a restricted form of orthonormal projection, since retaining $r$ original channels corresponds to choosing $r$ standard basis vectors.
An arbitrary projection therefore offers greater flexibility, but this flexibility alone does not guarantee more accurate token ranking.
The subspace must be chosen using an objective that preserves the query--key interactions that determine attention scores.
Figure~\ref{fig:motivation}(a) demonstrates this distinction: FASA's discrete channel selection outperforms Loki's key-reconstruction projection at every evaluated rank, with particularly large gaps at tight budgets.
The comparison motivates choosing the scoring subspace to preserve query--key interactions rather than key representations alone.

\subsection{Gather Overhead Limits Token-Level Speedup}
\label{sec:gather-overhead}

Even accurate token selection faces a separate systems challenge.
Fine-grained selection can produce irregular index sets that do not map naturally to the layouts supported by conventional attention kernels.
A common implementation therefore gathers the selected key-value entries into a contiguous buffer before invoking a dense attention kernel.
Figure~\ref{fig:motivation}(b) shows this overhead on LLaMA-3.1-8B-Instruct~\citep{llama3} at 128K context by comparing gather-then-attend with reduced attention over the same selected tokens when the KV entries are already contiguous.
Although sparsity substantially lowers total latency, gathering remains comparable to the reduced-attention computation itself and accounts for more than half of the residual latency at batch size 16.
Thus, physically materializing selected KV entries can erase a substantial fraction of the benefit enabled by token-level sparsity.

These observations motivate two complementary requirements: a low-dimensional scoring subspace that preserves token rankings, and an attention kernel that directly accesses the selected KV entries.

\section{Joint Subspace Sparse Attention}
\label{sec:method}

\subsection{Overview}
\label{sec:method_overview}

JSSA is a training-free token-level sparse-attention method for GQA and MLA, as illustrated in Figure~\ref{fig:overview}.
For each layer and KV head, it maintains a low-dimensional projected-key cache $\mathbf{K}_{\mathrm{proj},h}$ alongside the original KV cache.
At each decoding step, JSSA projects the mean query $\bar{\mathbf{q}}_{t,h}$ of the $G$ query heads associated with KV head $h$, scores it against $\mathbf{K}_{\mathrm{proj},h}$, and selects the top-$B$ token positions.
The associated query heads share this per-KV-head index set, while attention is computed from the original full-dimensional queries, keys, and values.
For GQA, our gather-free indexed kernel directly accesses the selected KV entries; for MLA, we use FlashMLA's sparse attention kernel~\citep{flashmla2025}.

\subsection{Joint Query--Key Subspace}
\label{sec:joint_subspace}

For each layer and KV head, JSSA uses an orthonormal basis
$M \in \mathbb{R}^{r \times d}$ with $MM^\top = I_r$, and applies the same basis to both queries and keys.
Let $P = M^\top M$ be the corresponding rank-$r$ projector.
Unlike asymmetric query--key factorizations such as KQ-SVD~\citep{kqsvd}, JSSA constrains both operands to a single shared scoring subspace; Appendix~\ref{app:kqsvd_comparison} evaluates an asymmetric KQ-SVD factorization adapted to the same token-selection pipeline.
In GQA, selection uses the mean query
$\bar{q} = \frac{1}{G}\sum_{g=1}^{G} q_g$
of the query heads associated with the KV head.
The projected score is
\begin{equation}
\hat{s} = \bar{q}^{\top}Pk,
\end{equation}
and the resulting score error is
\begin{equation}
s-\hat{s} = \bar{q}^{\top}(I-P)k.
\end{equation}

We therefore choose $P$ to minimize the expected squared error of the bilinear score used for token ranking.
Unlike query- or key-reconstruction objectives, this objective directly preserves the query--key interaction that determines token ordering.
For GQA, it is defined over the group-mean query used for per-KV-head selection.

Let $\widetilde{G}_{\bar{Q}}$ and $\widetilde{G}_{K}$ denote the independently unit-trace-normalized second-moment matrices of the calibration group-mean queries and keys, preventing either marginal from dominating the surrogate.
The exact selection objective is
\begin{equation}
\label{eq:selection_objective}
\mathcal{L}_{\mathrm{sel}}(P)
=
\operatorname{tr}
\left(
(I-P)\widetilde{G}_{\bar{Q}}(I-P)\widetilde{G}_{K}
\right).
\end{equation}

The exact objective is non-convex and has no general closed-form solution.
For any orthogonal projector $P$, however,
\begin{equation}
\mathcal{L}_{\mathrm{sel}}(P)
\le
\frac{1}{4}
\mathcal{L}_{\mathrm{marg}}(P)^2,
\qquad
\mathcal{L}_{\mathrm{marg}}(P)
=
\operatorname{tr}
\left(
(I-P)
\left(
\widetilde{G}_{\bar{Q}}+\widetilde{G}_{K}
\right)
\right).
\end{equation}
The full derivation is provided in Appendix~\ref{app:marginal_bound_derivation}.
The marginal objective admits the closed-form solution
\begin{equation}
M^\star
=
\operatorname{TopEig}_r
\left(
\widetilde{G}_{\bar{Q}}+\widetilde{G}_{K}
\right)^\top,
\end{equation}
where $\operatorname{TopEig}_r(A)$ returns the top-$r$ eigenvectors of $A$ as columns.
The bound provides a tractable surrogate but does not establish tightness or global optimality.
Direct optimization from $M^\star$ produces negligible subspace changes and no measurable improvement in oracle top-$B$ recall across the tested settings (Appendix~\ref{app:objective_refinement}).
We therefore use $M^\star$ directly, reducing calibration to second-moment accumulation followed by one eigendecomposition per layer and KV head.

\subsection{Gather-Free Indexed Attention}
\label{sec:gather_free}
After selecting the top-$B$ positions for each KV head, JSSA computes attention directly from their scattered locations in the original full-dimensional KV cache. Irregular indices are not themselves a barrier at this stage: a single token's key or value slice is already contiguous over head dim, so each indexed read remains a burst-sized transaction and irregularity only increases the number of transactions rather than degrading each one.
The main challenge in GQA lies instead in the cache layout, where each KV head occupies a separate partition and may select a different token set. Collapsing selections into a single layer-shared set reduces accuracy at long context (Appendix~\ref{app:gqa_selection_granularity}). This issue does not arise in MLA, where each token contributes one shared latent vector per layer and selection produces a single index set, the regime targeted by the released FlashMLA sparse kernels.

We therefore develop a Triton indexed-attention kernel that, to the best of our knowledge, is the first to support gather-free token-level attention with distinct per-KV-head selections under the standard GQA KV-cache layout. It launches once per layer and resolves the per-head index sets internally, mapping each query head to the indices selected by its associated KV head and accessing the corresponding keys and values in place. This avoids both the staging buffer required by gather-then-attend and the per-KV-head launches otherwise needed to keep distinct selections. 
Token scoring still examines all $N$ candidates, whereas attention accesses only the selected $B$ positions; Section~\ref{sec:ablations} shows consistent end-to-end speedups over gather-then-attend.

\subsection{Projected Scoring Cache}
\label{sec:scoring_cache}

JSSA maintains
$\mathbf{K}_{\mathrm{proj}}=\mathbf{K}M^\top$
as a scoring-only auxiliary cache for token selection, while final attention uses the original full-dimensional keys and values.
For GQA models, we choose $r$ so that the FP8 projected-key cache occupies at most $6.25\%$ of the original BF16 KV cache.
For MLA, we use a 128-dimensional scoring representation comprising 64 latent and 64 RoPE dimensions, matching the total dimensionality of the DSA indexer for an equal-capacity comparison.
We use FP8 throughout; Appendix~\ref{app:scoring_cache} verifies negligible accuracy impact relative to BF16.

\section{Experiments}
\label{sec:experiments}

\subsection{Setup}
\label{sec:setup}

We evaluate JSSA across standard GQA, hybrid local/global GQA, and MLA architectures.
LongBench~\citep{longbench}, RULER~\citep{ruler}, and PG-19~\citep{pg19} use LLaMA-3.1-8B-Instruct, Qwen3-14B~\citep{qwen3}, and Ministral-3-8B-Instruct-2512~\citep{ministral3}.
Reasoning and throughput experiments use Qwen3-8B, Gemma4-26B-A4B-It~\citep{gemma4}, and GLM-4.7-Flash~\citep{glm45}, respectively representing GQA, hybrid local/global GQA, and MLA.
For Gemma4, JSSA is applied only to the global-attention layers, while local layers retain the native sliding-window attention.
We compare against full-KV and the supported sparse-attention baselines for each architecture; complete model coverage, decoding configurations, and implementation details are provided in Appendix~\ref{app:setup}.
For GQA models, we set the FP8 projected-key cache to at most 6.25\% of the original BF16 KV cache ($r=32$ for LLaMA, Qwen, and Ministral; $r=64$ for Gemma4); for MLA, we use a 128-dimensional scoring representation matching the total dimensionality of the DSA indexer.

\subsection{Long-context Understanding}

\begin{table}[t]
\centering
\caption{
Average LongBench performance at a KV budget of 256 tokens.
FASA, Loki, and JSSA use 32-dimensional scoring representations.
The best and second-best sparse results for each model are shown in bold and underlined, respectively.
}
\vspace{-2pt}
\label{tab:longbench}
\renewcommand{\arraystretch}{1.05}
\footnotesize
\resizebox{\linewidth}{!}{
\begin{tabular}{lccccccc}
\toprule
Model & \cellcolor{gray!12}FKV & QUEST & SqueezedAttn. & SnapKV & FASA & Loki & \cellcolor{teal!10}JSSA \\
\midrule
LLaMA-3.1-8B-Instruct & \cellcolor{gray!12}48.35 & 46.94 & \underline{47.95} & 43.53 & 46.74 & 45.65 & \cellcolor{teal!10}\textbf{49.16} \\
Qwen3-14B & \cellcolor{gray!12}50.49 & 48.44 & 48.58 & 45.27 & 47.93 & \underline{49.54} & \cellcolor{teal!10}\textbf{49.82} \\
Ministral-3-8B-Instruct-2512 & \cellcolor{gray!12}47.90 & 46.48 & \underline{47.91} & 44.82 & 46.04 & \textbf{48.07} & \cellcolor{teal!10}47.79 \\
\bottomrule
\end{tabular}
}
\end{table}

\subsubsection{LongBench}
\label{sec:longbench}
Table~\ref{tab:longbench} reports average LongBench performance at a KV budget of 256 tokens. 
JSSA achieves the best or near-best sparse result on all three models and remains within 0.67 points of full-KV, while no single competing baseline is consistently strongest across architectures.

\begin{figure}[t]
\centering
\vspace{-8pt}
\includegraphics[width=1.0\textwidth]{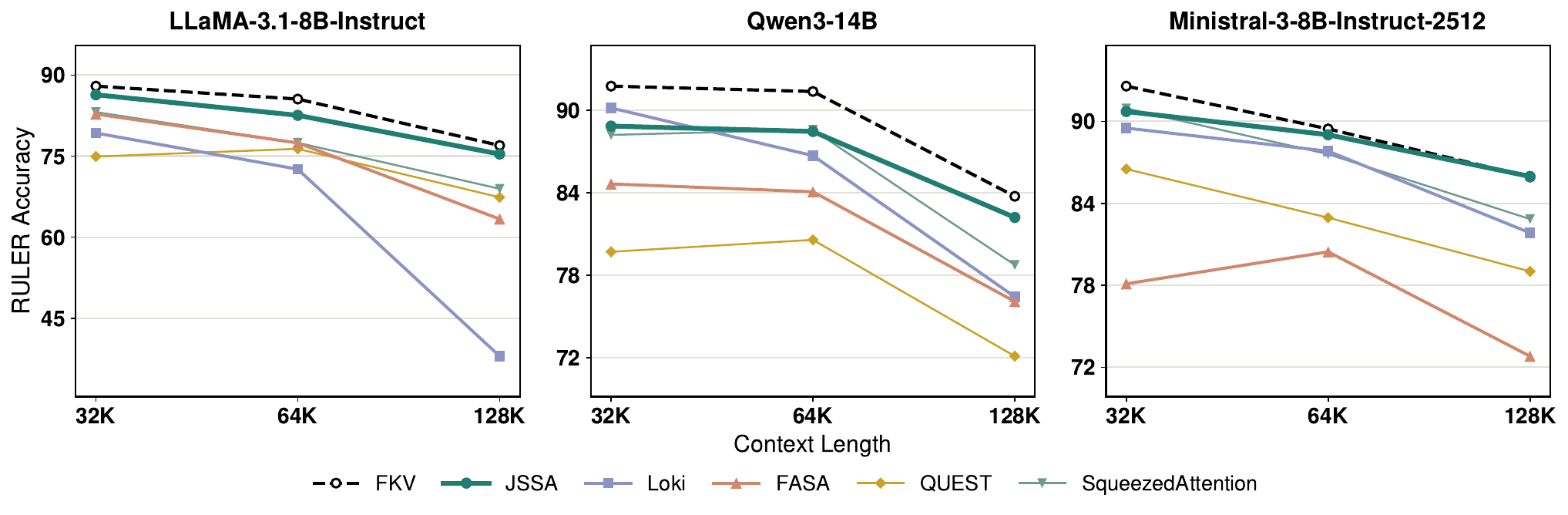}
\vspace{-16pt}
\caption{
\textbf{RULER accuracy across context lengths.}
Average accuracy under a fixed $1.5625\%$ KV budget, corresponding to 512, 1,024, and 2,048 selected tokens at 32K, 64K, and 128K context, respectively.
JSSA remains close to full-KV across models, while competing methods exhibit stronger model- and length-dependent degradation.
}
\label{fig:ruler}
\end{figure}

\subsubsection{RULER}
\label{sec:ruler}
Figure~\ref{fig:ruler} reports average RULER accuracy from 32K to 128K context under a fixed $1.5625\%$ KV budget. 
JSSA remains closest to full-KV attention overall and achieves the highest sparse accuracy at 128K, while competing methods show sharper degradation at long context (e.g., Loki on LLaMA-3.1-8B-Instruct) or model-specific weaknesses; on Ministral-3-8B-Instruct-2512, JSSA matches full-KV at 128K and is within 2 points at 32K and 64K.

\subsection{Language Modeling}
\label{sec:language_modeling}

Table~\ref{tab:pg19} reports PG-19 perplexity from 32K to 128K context under the same fixed-density budgets used for RULER. 
JSSA remains close to full-KV attention across all three models and context lengths, with a maximum gap of 1.11 perplexity points, whereas competing methods are far less consistent (e.g., Loki degrades sharply on LLaMA despite performing well on Ministral, and FASA's perplexity spikes on LLaMA at 128K).

\begin{table}[t]
\centering
\vspace{-6pt}
\caption{
PG-19 perplexity across context lengths under the same fixed-density KV budgets used for RULER.
Lower is better.
The best and second-best sparse results are shown in bold and underlined, respectively.
}
\label{tab:pg19}
\small
\setlength{\tabcolsep}{4.5pt}
\begin{tabular}{llccccc}
\toprule
Model & Context & \cellcolor{gray!12}FKV & QUEST & FASA & Loki & \cellcolor{teal!10}JSSA \\
\midrule
\multirow{3}{*}{LLaMA-3.1-8B-Instruct}
& 32K  & \cellcolor{gray!12}11.07 & 35.10 & \underline{23.12} & 50.15 & \cellcolor{teal!10}\textbf{11.27} \\
& 64K  & \cellcolor{gray!12}10.14 & 21.68 & \underline{14.76} & 41.45 & \cellcolor{teal!10}\textbf{10.14} \\
& 128K & \cellcolor{gray!12}10.38 & \underline{25.50} & 56.59 & 56.29 & \cellcolor{teal!10}\textbf{11.49} \\
\midrule
\multirow{3}{*}{Qwen3-14B}
& 32K  & \cellcolor{gray!12}11.24 & 16.38 & 16.89 & \underline{14.89} & \cellcolor{teal!10}\textbf{11.49} \\
& 64K  & \cellcolor{gray!12}10.93 & 14.51 & 19.34 & \underline{13.49} & \cellcolor{teal!10}\textbf{11.01} \\
& 128K & \cellcolor{gray!12}10.41 & \underline{13.29} & 15.05 & 14.61 & \cellcolor{teal!10}\textbf{10.53} \\
\midrule
\multirow{3}{*}{Ministral-3-8B-Instruct-2512}
& 32K  & \cellcolor{gray!12}9.78 & 13.87 & 24.13 & \textbf{10.43} & \cellcolor{teal!10}\underline{10.45} \\
& 64K  & \cellcolor{gray!12}8.96 & 12.46 & 16.20 & \textbf{9.55} & \cellcolor{teal!10}\underline{9.64} \\
& 128K & \cellcolor{gray!12}8.37 & 10.93 & 10.88 & \underline{9.46} & \cellcolor{teal!10}\textbf{9.05} \\
\bottomrule
\end{tabular}
\end{table}

\subsection{Reasoning}
\label{sec:reasoning}

Table~\ref{tab:reasoning} reports reasoning performance at a KV budget of 1,024 tokens across standard GQA, hybrid local/global GQA, and MLA models.
Each entry includes pass@1 accuracy and generation length relative to full-KV, since degraded sparse attention can destabilize reasoning trajectories even when the final answer remains correct.
Across architectures, JSSA provides the strongest overall accuracy--length trade-off, remaining within 1.3 points of full-KV accuracy on most benchmarks while limiting generation-length inflation to 2--17\% on Qwen3-8B and Gemma4-26B-A4B-It, versus substantially longer generations for TriAttention~\citep{triattention}, QUEST, and SeerAttention-R.
The gap over FASA is largest on Gemma4 (12.92 points on AIME 2025~\citep{aime25}) and GLM-4.7-Flash, where JSSA matches or exceeds full-KV accuracy on all three benchmarks (AIME 2025, MATH-500~\citep{math500}, and GPQA-Diamond~\citep{gpqa}).
Complete results across budgets of 512, 1,024, and 2,048 tokens are provided in Appendix~\ref{app:reasoning_results}.

\begin{table*}[t]
\centering
\vspace{-12pt}
\caption{
Reasoning performance at a KV budget of 1,024 tokens.
Each entry reports pass@1 accuracy, with generation length relative to FKV shown in parentheses.
Higher accuracy and lower length ratio are better.
Bold indicates the best sparse accuracy for each model and benchmark.
}
\vspace{0pt}
\label{tab:reasoning}
\renewcommand{\arraystretch}{0.95}
\scriptsize
\resizebox{\linewidth}{!}{
\begin{tabular}{llccc}
\toprule
Model & Method & AIME 2025 & MATH-500 & GPQA-Diamond \\
\midrule

\multirow{6}{*}{Qwen3-8B}
& \cellcolor{gray!12}FKV
& \cellcolor{gray!12}69.58 (1.00$\times$)
& \cellcolor{gray!12}94.53 (1.00$\times$)
& \cellcolor{gray!12}57.95 (1.00$\times$) \\
& TriAttention
& 20.00 (1.75$\times$)
& 86.20 (1.70$\times$)
& 40.40 (4.13$\times$) \\
& QUEST
& 62.92 (1.29$\times$)
& 93.50 (1.21$\times$)
& \textbf{58.14} (1.03$\times$) \\
& SeerAttention-R
& 42.08 (1.54$\times$)
& 91.80 (1.35$\times$)
& 55.93 (1.43$\times$) \\
& FASA
& 60.42 (1.30$\times$)
& 93.58 (1.17$\times$)
& 57.45 (1.06$\times$) \\
& \cellcolor{teal!10}JSSA
& \cellcolor{teal!10}\textbf{68.33} (1.07$\times$)
& \cellcolor{teal!10}\textbf{94.23} (1.09$\times$)
& \cellcolor{teal!10}57.39 (1.02$\times$) \\

\midrule

\multirow{5}{*}{Gemma4-26B-A4B-It}
& \cellcolor{gray!12}FKV
& \cellcolor{gray!12}90.00 (1.00$\times$)
& \cellcolor{gray!12}97.15 (1.00$\times$)
& \cellcolor{gray!12}80.87 (1.00$\times$) \\
& TriAttention
& 22.50 (2.29$\times$)
& 83.95 (2.37$\times$)
& 43.94 (3.06$\times$) \\
& QUEST
& 50.42 (1.96$\times$)
& 93.65 (1.65$\times$)
& 71.40 (2.02$\times$) \\
& FASA
& 73.75 (1.38$\times$)
& \textbf{97.08} (1.14$\times$)
& 80.56 (1.18$\times$) \\
& \cellcolor{teal!10}JSSA
& \cellcolor{teal!10}\textbf{86.67} (1.17$\times$)
& \cellcolor{teal!10}\textbf{97.08} (1.13$\times$)
& \cellcolor{teal!10}\textbf{81.94} (1.12$\times$) \\

\midrule

\multirow{3}{*}{GLM-4.7-Flash}
& \cellcolor{gray!12}FKV
& \cellcolor{gray!12}90.42 (1.00$\times$)
& \cellcolor{gray!12}96.80 (1.00$\times$)
& \cellcolor{gray!12}75.44 (1.00$\times$) \\
& FASA
& 65.00 (2.24$\times$)
& 95.65 (1.90$\times$)
& 69.38 (2.30$\times$) \\
& \cellcolor{teal!10}JSSA
& \cellcolor{teal!10}\textbf{91.67} (1.17$\times$)
& \cellcolor{teal!10}\textbf{96.88} (1.10$\times$)
& \cellcolor{teal!10}\textbf{77.53} (1.38$\times$) \\

\bottomrule
\end{tabular}
}
\end{table*}

\begin{figure}[t]
\centering
\vspace{-2pt}
\includegraphics[width=1.0\textwidth]{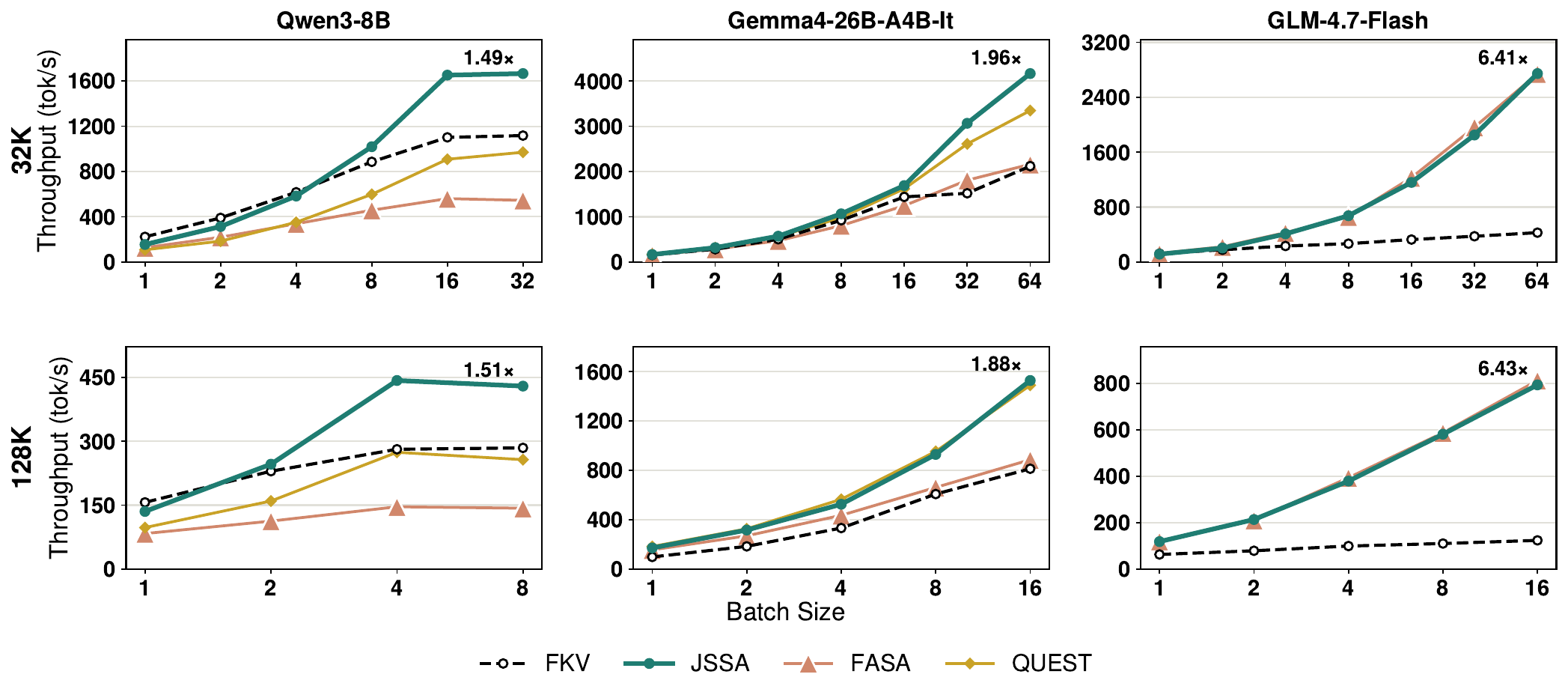}
\vspace{-16pt}
\caption{
\textbf{End-to-end decoding throughput.}
Throughput on a single NVIDIA B200 GPU at 32K and 128K context with a KV budget of $B=2{,}048$ tokens.
Annotations report JSSA's maximum speedup over FKV for each model and context setting.
}
\vspace{-4pt}
\label{fig:throughput}
\end{figure}

\subsection{Decoding Throughput}
\label{sec:throughput}

We measure end-to-end decoding throughput on a single NVIDIA B200 GPU across Qwen3-8B, Gemma4-26B-A4B-It, and GLM-4.7-Flash, covering GQA, hybrid sliding-window/global attention, and MLA architectures.
All sparse methods use a KV budget of $B=2{,}048$ tokens.
We sweep batch size at 32K and 128K context using the FP8 projected scoring cache and the gather-free indexed-attention kernel described in Sections~\ref{sec:scoring_cache} and~\ref{sec:gather_free}.

Figure~\ref{fig:throughput} reports the resulting throughput.
JSSA consistently improves over full-KV, and its advantage generally grows at 128K as KV-cache traffic becomes more dominant.
At 128K, JSSA reaches up to $1.51\times$, $1.88\times$, and $6.43\times$ the throughput of full-KV on Qwen3-8B, Gemma4-26B-A4B-It, and GLM-4.7-Flash, respectively.
On GLM-4.7-Flash, JSSA matches FASA's throughput while providing substantially higher reasoning accuracy, indicating that the joint scoring objective adds little overhead (complete sweeps in Appendix~\ref{app:throughput_details}).

\subsection{Validation at Large Scale}
\label{sec:large_scale}

We evaluate JSSA on two model families at 700B-parameter scale, DeepSeek-V3 and GLM-5.2, equipped with model-native DeepSeek Sparse Attention (DSA). 
This allows us to compare post-hoc JSSA calibration against sparse-attention mechanisms introduced during model training. 
For DeepSeek, we apply JSSA to DeepSeek-V3.1-Terminus (full attention) and compare against the officially released DeepSeek-V3.2-Exp, which introduces native DSA through continued pre-training from
the Terminus. 
For GLM-5.2, an independently developed MLA model, JSSA follows the model's native schedule of invoking selection only at a subset of layers and reusing indices across adjacent layers; no full-attention checkpoint is available for this model. 
Complete experimental details are provided in Appendix~\ref{app:large_scale}.

\begin{table}[h]
\centering
\vspace{-12pt}
\caption{
Large-scale reasoning accuracy.
JSSA is applied post hoc without model training.
FKV and JSSA use DeepSeek-V3.1-Terminus, whereas native DSA uses
DeepSeek-V3.2-Exp. No corresponding FKV checkpoint is available for GLM-5.2.
}
\vspace{-2pt}
\label{tab:large_scale_reasoning}
\renewcommand{\arraystretch}{1.0}
\setlength{\tabcolsep}{12pt}
\scalebox{0.95}{
\begin{tabular}{llcc}
\toprule
Model & Method & AIME 2025 & GPQA-Diamond \\
\midrule
DeepSeek-V3.1-Terminus & FKV & 86.67 & \textbf{81.57} \\
DeepSeek-V3.1-Terminus & JSSA & \textbf{87.50} & 80.30 \\
DeepSeek-V3.2-Exp & Native DSA & \textbf{87.50} & 78.54 \\
\midrule
GLM-5.2 & JSSA & \textbf{83.33} & \textbf{90.91} \\
GLM-5.2 & Native DSA & 82.50 & \textbf{90.91} \\
\bottomrule
\end{tabular}
}
\end{table}

Table~\ref{tab:large_scale_reasoning} reports pass@1 accuracy on AIME 2025 and GPQA-Diamond. Across both model families, JSSA, FKV, and native DSA remain within a narrow accuracy range: JSSA matches native DSA on AIME 2025 for both DeepSeek and GLM-5.2, and the two methods are identical on GPQA-Diamond for GLM-5.2. 
Given the small numerical differences involved, we interpret these results as showing accuracy comparable to model-native DSA rather than a consistent advantage for either method.
On throughput, JSSA reaches 96.1--99.2\% of native DSA's decoding throughput across the measured context lengths and batch sizes on DeepSeek, without requiring continued training. 
At 128K context, JSSA reaches up to $3.49\times$ the throughput of full-KV attention at batch size 8; complete sweeps are provided in Appendix~\ref{app:large_scale}.

\subsection{Ablation Studies}
\label{sec:ablations}

\begin{figure}[t]
\centering
\vspace{-4pt}
\includegraphics[width=1.0\textwidth]{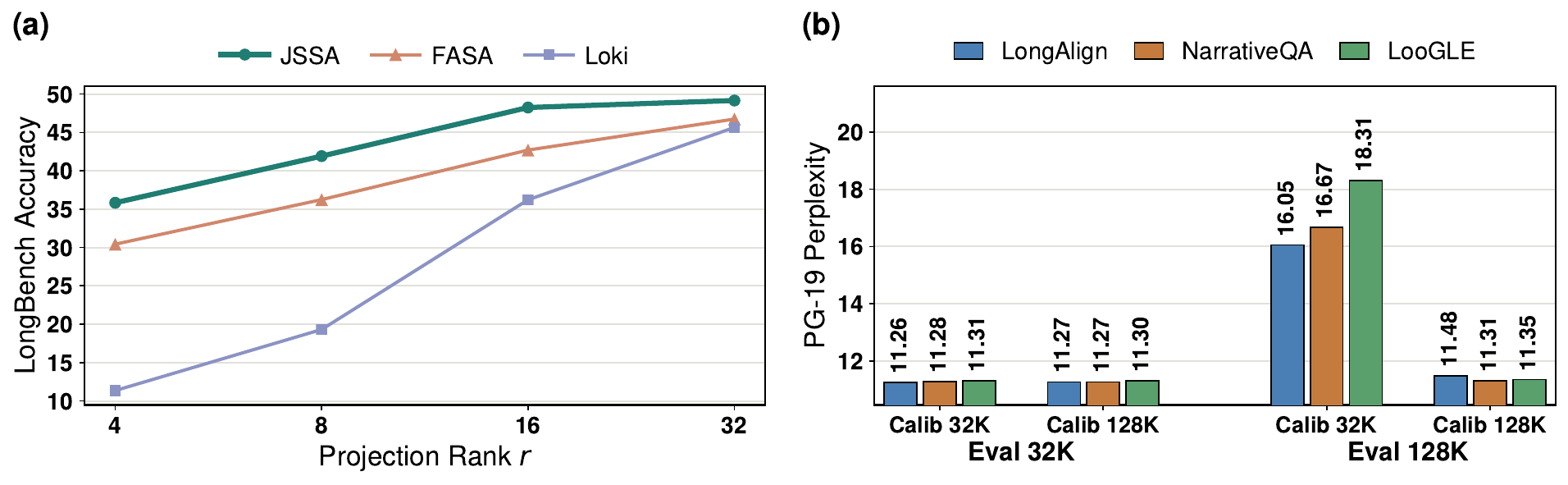}
\vspace{-16pt}
\caption{
\textbf{Scoring-subspace ablations.}
(a) LongBench accuracy versus projection rank on LLaMA-3.1-8B-Instruct at a KV budget of 256 tokens.
(b) PG-19 perplexity on LLaMA-3.1-8B-Instruct under the same fixed-density KV budgets used in the main evaluation, for bases calibrated on different corpora and context lengths.
Calibration transfers across datasets, but bases calibrated at 32K degrade when evaluated at 128K.
}
\vspace{-4pt}
\label{fig:scoring_ablations}
\end{figure}

\paragraph{Projection rank.}
Figure~\ref{fig:scoring_ablations}(a) compares JSSA with FASA and Loki as the scoring dimension varies from 4 to 32. JSSA performs best at every rank, with the largest gains under the tightest constraints (e.g., $+5.40$ and $+24.48$ LongBench points over FASA and Loki at $r=4$), showing that the joint query--key objective uses limited scoring capacity more effectively.

\paragraph{Calibration robustness.}
Figure~\ref{fig:scoring_ablations}(b) separates the effects of calibration corpus and calibration length under the same fixed-density KV budgets used in the main evaluation.
JSSA is largely insensitive to the calibration corpus: bases calibrated on LongAlign, NarrativeQA, and LooGLE yield nearly identical perplexity at both 32K and 128K evaluation.
Calibration length matters more: bases calibrated at 32K degrade sharply when evaluated at 128K (perplexity 16.05--18.31, versus 11.31--11.48 for 128K-calibrated bases), indicating that calibration should cover sequences near the maximum deployment context length.

\begin{figure}[t]
\centering
\vspace{-4pt}
\includegraphics[width=1.0\textwidth]{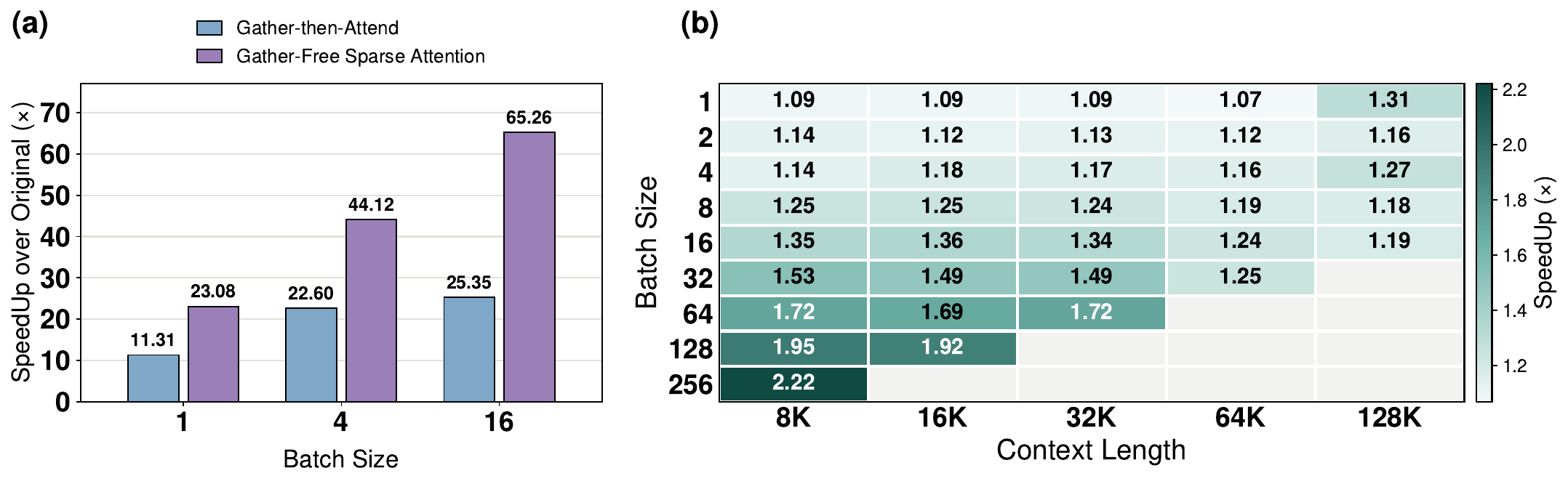}
\vspace{-18pt}
\caption{
\textbf{Effect of gather-free indexed attention} ($B=2{,}048$ tokens).
(a) Attention-stage speedup over full-KV attention on LLaMA-3.1-8B-Instruct at 128K, comparing gather-then-attend with our gather-free kernel.
(b) End-to-end decoding speedup of gather-free execution over gather-then-attend on Gemma4-26B-A4B-It across context lengths and batch sizes.
}
\label{fig:gather_free_ablations}
\end{figure}

\paragraph{Gather-free execution.}
Figure~\ref{fig:gather_free_ablations} isolates the benefit of consuming the selected KV entries directly through indexed attention rather than first materializing them into a contiguous buffer, using the same selected token sets with a KV budget of $B=2{,}048$ tokens. 
At 128K context on LLaMA-3.1-8B-Instruct, gather-free execution roughly doubles the attention-stage speedup over full-KV attention at batch sizes 4 and 16 (e.g., $25.35\times \to 65.26\times$ at batch 16). 
On Gemma4-26B-A4B-It, it improves end-to-end throughput by $1.07\times$--$2.22\times$ over gather-then-attend, with the gain growing with batch size, showing that physically gathering selected KV entries remains a substantial overhead even after sparsification.

\section{Related Work}
\label{sec:related_work}

\paragraph{Coarse-grained sparse attention.}
Coarse-grained methods amortize selection and execution over groups of tokens.
Page- and block-level approaches like QUEST~\citep{quest} and SeerAttention-R~\citep{seerattentionr} estimate query-dependent importance for KV pages that map naturally to block-sparse kernels.
For prefill, MInference~\citep{minference}, FlexPrefill~\citep{flexprefill}, and XAttention~\citep{xattention} identify input-dependent sparse patterns to reduce prompt-processing cost.
Model-native designs such as NSA~\citep{nsa} and MSA~\citep{msa2026} jointly train sparse selection and hardware-aligned execution across prefill and decoding.
SqueezedAttention~\citep{squeezedattention} instead retrieves representation-similar clusters via centroids. Because block- and cluster-level methods may access irrelevant members within selected groups, JSSA performs individual-token selection instead.

\paragraph{Token-level sparse attention and execution.}
Dynamic token-level methods reconsider individual KV positions for each decoding query.
SparQ~\citep{sparq} uses salient query channels, TokenSelect~\citep{tokenselect} aggregates head-specific token preferences, Loki~\citep{loki} scores tokens in a key-reconstruction subspace, and FASA~\citep{fasa} selects original RoPE channels.
SALS~\citep{sals} retrieves tokens from a compressed latent KV cache, while model-native DSA~\cite{deepseekv32} learns a low-dimensional token indexer during training.
Related KV-reduction methods, including H$_2$O~\citep{h2o}, SnapKV~\citep{snapkv}, and TriAttention~\citep{triattention}, permanently evict token-level entries rather than dynamically retrieving from the complete cache.
JSSA retains the complete cache, derives its scoring space from joint query--key score-preservation error, and preserves distinct selected sets across KV heads.
Existing efficient execution paths primarily target regular block selections or MLA's shared latent cache, including the sparse kernels released in FlashMLA~\citep{flashmla2025}.
JSSA instead provides gather-free indexed attention for irregular per-KV-head selections under the standard GQA KV-cache layout.

\paragraph{Projected and low-rank attention representations.}
Low-dimensional transformations have also been used to reduce KV storage or final-attention computation.
EigenAttention~\citep{eigenattention} evaluates attention in a reduced-dimensional space, KVTC~\citep{kvtc} applies a decorrelating transform before compression, and SWAN~\citep{swan} rotates and prunes KV features using an orthogonal transformation.
KQ-SVD~\citep{kqsvd} instead derives asymmetric query and key projections that directly target attention-score fidelity; Section~\ref{sec:method} and Appendix~\ref{app:kqsvd_comparison} compare this asymmetric parameterization with JSSA's shared scoring subspace.
These methods replace or compress the representation consumed by final attention, whereas JSSA retains the original full-dimensional KV cache and uses projection only for an auxiliary scoring cache.

\section{Conclusion}

We presented JSSA, a training-free token-level sparse-attention method that jointly addresses selection quality and execution efficiency in long-context decoding.
JSSA derives a shared query--key scoring subspace from a closed-form spectral objective and eliminates token-gathering overhead through gather-free indexed attention with per-KV-head selections in GQA.
Across GQA, hybrid local/global attention, and MLA models, JSSA remains close to full-KV accuracy while delivering substantial throughput gains.
At approximately 700B scale, JSSA matches model-native DSA's reasoning accuracy while reaching 96–99\% of its decoding throughput, without model training.

\bibliography{iclr2027_conference}
\bibliographystyle{iclr2027_conference}

\appendix

\section{Setup Details}
\label{app:setup}

\subsection{Models and Evaluation Coverage}
\label{app:model_coverage}

Table~\ref{tab:model_coverage} summarizes the models and evaluations used in this work.
LongBench, RULER, and PG-19 use LLaMA-3.1-8B-Instruct, Qwen3-14B, and Ministral-3-8B-Instruct-2512, enabling direct comparison of sparse-selection behavior across long-context understanding and language modeling.
Reasoning and throughput experiments use Qwen3-8B, Gemma4-26B-A4B-It, and GLM-4.7-Flash as representative models with standard GQA, hybrid local/global GQA, and MLA, respectively.

For Gemma4, JSSA is applied only to the global-attention layers, while the local layers retain the model's native sliding-window attention.
For GLM-4.7-Flash, token selection operates over the shared MLA cache, and sparse attention is executed using FlashMLA.

\begin{table}[h]
\centering
\caption{
\textbf{Model and evaluation coverage.}
Checkmarks indicate the evaluations performed for each model.
For Gemma4, JSSA is applied only to global-attention layers, while local sliding-window layers remain unchanged.
}
\vspace{4pt}
\label{tab:model_coverage}
\renewcommand{\arraystretch}{1.05}
\setlength{\tabcolsep}{2pt}
\scalebox{0.9}{
\begin{tabular}{llccccc}
\toprule
Model & Attention architecture & LongBench & RULER & PG-19 & Reasoning & Throughput \\
\midrule
LLaMA-3.1-8B-Instruct & GQA & \checkmark & \checkmark & \checkmark & & \\
Qwen3-14B & GQA & \checkmark & \checkmark & \checkmark & & \\
Ministral-3-8B-Instruct-2512 & GQA & \checkmark & \checkmark & \checkmark & & \\
\midrule
Qwen3-8B & GQA & & & & \checkmark & \checkmark \\
Gemma4-26B-A4B-It & Hybrid local/global GQA & & & & \checkmark & \checkmark \\
GLM-4.7-Flash & MLA & & & & \checkmark & \checkmark \\
\bottomrule
\end{tabular}
}
\end{table}

\subsection{Evaluation Protocols}
\label{app:evaluation_protocols}

For LongBench and RULER, we use the benchmark-provided prompts, task-specific generation limits, and official evaluation scripts with greedy decoding.
PG-19 is evaluated using teacher-forced perplexity.
For reasoning benchmarks, we follow each model's recommended sampling configuration and use the benchmark-specific generation limits and scoring procedures.
All compared methods use identical decoding parameters, evaluation examples, and random seeds within each model and benchmark.
Each reasoning evaluation is repeated with eight independent seeds, and we report the mean pass@1 accuracy and generation length.

Table~\ref{tab:reasoning_decoding_config} reports the model-specific decoding parameters used for reasoning evaluations.

\begin{table}[h]
\centering
\caption{
\textbf{Reasoning decoding configurations.}
All compared methods use identical parameters and evaluation seeds for a given model and benchmark.
Each evaluation is repeated with eight independent seeds, and the reported results are averaged across runs.
}
\vspace{4pt}
\label{tab:reasoning_decoding_config}
\renewcommand{\arraystretch}{1.05}
\setlength{\tabcolsep}{8pt}
\footnotesize
\begin{tabular}{lcccc}
\toprule
Model & Temperature & Top-$p$ & Top-$k$ & Max. new tokens \\
\midrule
Qwen3-8B & 0.6 & 0.95 & 20 & 40960 \\
Gemma4-26B-A4B-It & 1.0 & 0.95 & 64 & 40960 \\
GLM-4.7-Flash & 1.0 & 0.95 & - & 131072 \\
\bottomrule
\end{tabular}
\end{table}

\subsection{JSSA Configuration}
\label{app:jssa_config}

For GQA models, JSSA maintains one projected-key scoring cache for each layer and KV head.
The query heads associated with a KV head share the token set selected using their mean projected query.
We use $r=32$ for LLaMA-3.1-8B-Instruct, Qwen3-8B, Qwen3-14B, and Ministral-3-8B-Instruct-2512, and $r=64$ for the global-attention layers of Gemma4-26B-A4B-It.
With FP8 storage, these projected-key caches occupy at most $6.25\%$ of the corresponding BF16 KV caches.

For MLA models, JSSA uses a 128-dimensional scoring representation comprising 64 latent and 64 RoPE dimensions, matching the total dimensionality used by DSA.
Selection operates over the shared latent cache and therefore produces one layer-level token set rather than distinct per-KV-head selections.
Sparse attention is executed using FlashMLA.

Across all architectures, the projected scoring cache is stored in FP8, while final sparse attention uses the original full-dimensional BF16 KV representations.
We apply the same token budget to every sparse-attention layer and head.
JSSA does not reserve sink tokens, enforce a recent-token window, or apply special handling to any token positions.
For Gemma4, only the global-attention layers are sparsified; the local layers retain the model's native sliding-window attention.
Appendix~\ref{app:scoring_cache} evaluates the accuracy impact of scoring-cache quantization.

\subsection{JSSA Calibration}
\label{app:jssa_calibration}

JSSA is calibrated without updating any model parameters.
Because individual LongAlign-10K samples extend only to approximately 64K tokens, using them directly would not expose the projection bases to the 128K-context attention patterns targeted in our evaluation.
We therefore filter the corpus to documents containing at least 8K tokens, concatenate consecutive documents in corpus order until reaching at least 150K tokens, and center-crop the result to 131,072 tokens.
We construct eight calibration sequences using starting offsets separated by five documents; the resulting sequences may overlap because each concatenated window can span more than five documents.

For each sequence, we run a dense prefill pass and accumulate the second moments of group-mean queries and keys separately for every layer and KV head.
We then compute one projection basis per layer and KV head using the closed-form solution in Equation~\ref{eq:app_closed_form_basis}.
For Gemma4, calibration is performed only for the global-attention layers, while the local sliding-window layers remain unchanged.
No gradient-based optimization, model training, or downstream evaluation data is used.

\paragraph{Calibration cost.}
Table~\ref{tab:calibration_latency} reports the end-to-end wall-clock time required to construct all JSSA bases.
Calibration completes within 129--365 seconds across the evaluated models and requires neither iterative optimization nor model training.

\begin{table}[h]
\centering
\caption{End-to-end JSSA calibration latency on an NVIDIA B200 GPU.}
\label{tab:calibration_latency}
\small
\setlength{\tabcolsep}{4pt}
\begin{tabular}{lccccc}
\toprule
Model & LLaMA-3.1-8B & Qwen3-14B & Ministral-3-8B & Gemma4-26B & GLM-4.7-Flash \\
\midrule
Latency (s) & 203 & 295 & 129 & 365 & 247 \\
\bottomrule
\end{tabular}
\end{table}

\subsection{Calibration Input Construction}
\label{app:calibration_construction}

The main evaluation uses LongAlign as the common calibration corpus for all calibration-dependent methods, but the construction of calibration inputs is chosen separately for each method.
As shown in Section~\ref{sec:ablations}, JSSA requires calibration sequences that cover the intended deployment context range.
We therefore use the packed 128K-token LongAlign sequences described in Appendix~\ref{app:jssa_calibration} for JSSA.
For FASA, we compare the same packed construction against calibration on the original unmodified examples.
Loki is insensitive to this choice and uses the packed LongAlign calibration set.

\begin{table}[h]
\centering
\caption{
Average LongBench accuracy under different calibration corpora and input constructions.
Packed calibration uses 128K-token sequences, whereas plain calibration uses the original unmodified examples.
}
\label{tab:calibration_construction}
\vspace{4pt}
\renewcommand{\arraystretch}{1.05}
\setlength{\tabcolsep}{10pt}
\begin{tabular}{llcc}
\toprule
Method & Calibration Corpus & Packed 128K & Plain \\
\midrule
\multirow{2}{*}{JSSA}
& LongAlign & 49.16 & 49.06 \\
& Qasper    & 49.13 & 49.15 \\
\midrule
\multirow{2}{*}{FASA}
& LongAlign & 43.99 & 46.74 \\
& Qasper    & 44.70 & 47.21 \\
\bottomrule
\end{tabular}
\end{table}

Table~\ref{tab:calibration_construction} shows that JSSA is largely insensitive to both calibration corpus and input construction on LongBench, with accuracy varying by at most 0.10 points.
FASA is substantially more sensitive to input construction.
Replacing packed sequences with plain examples improves accuracy by 2.75 points with LongAlign and 2.51 points with Qasper.
By comparison, changing the calibration corpus within either construction has a smaller effect.
This pattern indicates that FASA's degradation under packed calibration is not specific to a particular corpus.

Based on these results, the main evaluation uses a fixed method-specific construction.
JSSA is calibrated on packed 128K-token LongAlign sequences so that calibration covers the deployment context range, whereas FASA is calibrated on plain LongAlign examples.
Loki uses the packed LongAlign calibration set because its results are insensitive to packing.
These settings are fixed across RULER, LongBench, language modeling, and reasoning evaluations, without benchmark-specific recalibration or tuning.

\subsection{Baseline Implementations}
\label{app:baseline_implementations}

We evaluate Loki, FASA, and QUEST within the same vLLM-based serving stack as JSSA.
SqueezedAttention, SnapKV, SeerAttention-R, and TriAttention are integrated into Hugging Face Transformers from their official codebases.
We evaluate each baseline only where its original formulation can be transferred without substantial method-specific design choices.

\paragraph{Loki.}
Loki shares JSSA's projected scoring cache, top-$B$ selection pipeline, and gather-free indexed-attention backend, differing only in its key-reconstruction projection objective.
We match its projection rank, cache precision, token budget, and execution path to JSSA, isolating the effect of the projection objective.

\paragraph{FASA.}
We integrate FASA's discrete channel-selection and scoring procedure into our vLLM-based serving stack.
Because FASA does not specify the selection granularity for GQA, we perform scoring and token selection independently for each KV head using the mean of its associated query heads.
We use the same scoring dimension and token budget as JSSA.
For each model, FASA is calibrated once on the original, unmodified LongAlign examples, and the resulting calibration is reused across benchmarks and context lengths without benchmark-specific recalibration or tuning.
Appendix~\ref{app:calibration_construction} compares plain and packed calibration inputs and motivates this fixed choice.
Following FASA's deployment design, selected channels are read directly from the original key cache rather than stored in a separate projected scoring cache.

\paragraph{QUEST.}
We adapt QUEST from MHA to GQA by performing page scoring and selection independently for each KV head.
The open page containing the most recent tokens is always retained, following QUEST's native paged-cache behavior.
To preserve vLLM's standard KV-cache layout while supporting distinct selected page sets, we launch the sparse-attention kernel separately for each KV head.

\paragraph{MLA baselines.}
Most existing baselines do not define a direct mapping to MLA's shared latent cache, and adapting them would require additional choices in scoring representation and selection semantics.
We therefore implement only FASA as a representative post-hoc baseline, applying it to all 64 channels of the RoPE cache.
This already uses half of the 128-dimensional scoring representation used by JSSA and DSA.

\paragraph{Other baselines.}
SqueezedAttention and SeerAttention-R retain their official coarse-grained selection procedures and recommended model-specific configurations.
SeerAttention-R additionally uses its released model-specific trained checkpoint and is therefore evaluated only on Qwen3-8B.
SnapKV and TriAttention perform token-level KV eviction rather than per-step dynamic retrieval; we retain their official eviction procedures and recommended configurations.

\subsection{Hardware and Measurement Protocol}
\label{app:hardware_protocol}

Unless otherwise stated, system experiments are conducted on a single NVIDIA B200 GPU with tensor parallelism degree one using our vLLM-based implementation.
CUDA graph capture is enabled, and each run generates 128 new tokens from the target input context.
For each configuration, we perform one warm-up run followed by three measured runs and report the mean aggregate decoding throughput.

Measurements include projected-score computation, top-$B$ selection, sparse attention, and the remaining model execution during decoding.
They exclude model loading, JSSA calibration, kernel compilation, and warm-up.
The original full-dimensional KV cache is stored in BF16, while JSSA's projected scoring cache is stored in FP8.

\section{Joint Subspace Objective Details}
\label{app:subspace_details}

\subsection{Product-of-Marginals Formulation}
\label{app:product_marginals}

Let $\mathcal{D}_{\bar Q}$ and $\mathcal{D}_K$ denote the empirical marginals of group-mean queries and keys collected during calibration.
We define the selection objective over independent draws
\begin{equation}
    \bar q \sim \mathcal{D}_{\bar Q},
    \qquad
    k \sim \mathcal{D}_K,
\end{equation}
or equivalently over the product distribution $\mathcal{D}_{\bar Q}\times\mathcal{D}_K$.

This formulation reflects the runtime use of the selector, where each decoding query is scored against every key in the context rather than only against the key observed at the same sequence position.
The calibration objective therefore measures the projected-score error $\bar q^\top(I-P)k$ across the query--key combinations that may be compared during inference.
Appendix~\ref{app:trace_normalization} defines the exact selection objective as the second moment of this error after trace normalization.

Because query--key pairs are drawn from the product distribution, $\bar q$ and $k$ are independent by construction.
Thus, for any functions $f$ and $g$,
\begin{equation}
    \mathbb{E}
    \left[
        f(\bar q)g(k)
    \right]
    =
    \mathbb{E}
    \left[
        f(\bar q)
    \right]
    \mathbb{E}
    \left[
        g(k)
    \right].
    \label{eq:app_product_factorization}
\end{equation}
Since trace normalization rescales each marginal by a constant, the same factorization holds for the rescaled activations in Appendix~\ref{app:trace_normalization}.

This factorization is used in the derivation of the marginal bound in Appendix~\ref{app:marginal_bound_derivation}.
It does not assert that query and key activations observed at corresponding sequence positions are empirically independent.

\subsection{Trace Normalization}
\label{app:trace_normalization}

Let
\begin{equation}
    G_{\bar Q}
    =
    \mathbb{E}_{\bar q\sim\mathcal D_{\bar Q}}
    \left[
        \bar q\bar q^\top
    \right],
    \qquad
    G_K
    =
    \mathbb{E}_{k\sim\mathcal D_K}
    \left[
        kk^\top
    \right]
\end{equation}
denote the second-moment matrices of the group-mean queries and keys.

We normalize the two marginals independently:
\begin{equation}
    \widetilde G_{\bar Q}
    =
    \frac{G_{\bar Q}}
         {\operatorname{tr}(G_{\bar Q})},
    \qquad
    \widetilde G_K
    =
    \frac{G_K}
         {\operatorname{tr}(G_K)}.
    \label{eq:app_normalized_grams}
\end{equation}

Equivalently, define
\begin{equation}
    \widetilde q
    =
    \frac{\bar q}
         {\sqrt{\operatorname{tr}(G_{\bar Q})}},
    \qquad
    \widetilde k
    =
    \frac{k}
         {\sqrt{\operatorname{tr}(G_K)}}.
    \label{eq:app_rescaled_activations}
\end{equation}

For a projector $P$, the exact score-error objective is
\begin{align}
    L_{\mathrm{sel}}(P)
    &=
    \mathbb{E}_{\substack{
        \bar q\sim\mathcal{D}_{\bar Q}\\
        k\sim\mathcal{D}_K
    }}
    \left[
        \left(
            \widetilde q^\top(I-P)\widetilde k
        \right)^2
    \right]
    \nonumber\\
    &=
    \operatorname{tr}
    \left(
        (I-P)\widetilde G_{\bar Q}(I-P)\widetilde G_K
    \right),
    \label{eq:app_all_pairs_objective}
\end{align}
which is the objective in Equation~\ref{eq:selection_objective} of the main text.
It differs from the second moment of the unnormalized score error only by a constant factor:
\begin{equation}
    L_{\mathrm{sel}}(P)
    =
    \frac{
        \mathbb{E}
        \left[
            \left(
                \bar q^\top(I-P)k
            \right)^2
        \right]
    }{
        \operatorname{tr}(G_{\bar Q})
        \operatorname{tr}(G_K)
    }.
    \label{eq:app_normalized_exact_objective}
\end{equation}

Because the scaling factor is positive and independent of \(P\), trace normalization does not change the minimizer of the exact selection objective.
It does, however, determine the relative weighting of the query and key marginals in the tractable surrogate.
For group-mean queries, we measure an average trace ratio of
\(\operatorname{tr}(G_{\bar Q})/\operatorname{tr}(G_K)=0.323\) on LLaMA-3.1-8B-Instruct.
Thus, the unnormalized sum \(G_{\bar Q}+G_K\) is strongly key-dominated and shifts the resulting basis toward a key-reconstruction-like subspace.
Independent unit-trace normalization instead assigns equal total weight to the query and key marginals while preserving their internal spectral structure.

\begin{table}[h]
\centering
\caption{
Effect of trace normalization on oracle top-\(B\) recall for LLaMA-3.1-8B-Instruct at \(r=32\).
}
\label{tab:trace_normalization}
\begin{tabular}{lcc}
\toprule
Basis & \(B=256\) & \(B=2048\) \\
\midrule
Raw group-mean moments & 70.05\% & 74.92\% \\
Unit-trace-normalized group-mean moments & 71.34\% & 75.94\% \\
\bottomrule
\end{tabular}
\end{table}

Table~\ref{tab:trace_normalization} shows that normalization improves oracle recall by 1.29 and 1.02 percentage points at token budgets of 256 and 2,048, respectively.
Although the absolute gains are modest, normalization prevents the marginal surrogate from being dominated by the larger key second moment and preserves its intended joint query--key character.

\subsection{Derivation of the Marginal Bound}
\label{app:marginal_bound_derivation}

Recall the exact selection objective
\begin{equation}
    L_{\mathrm{sel}}(P)
    =
    \mathbb{E}
    \left[
        \left(
            \widetilde q^\top(I-P)\widetilde k
        \right)^2
    \right],
\end{equation}
where the expectation is taken over the product distribution induced by the rescaled query and key marginals in Equation~\ref{eq:app_rescaled_activations}.

For any orthogonal projector $P$, define
\begin{equation}
    X
    =
    \widetilde q^\top(I-P)\widetilde q,
    \qquad
    Y
    =
    \widetilde k^\top(I-P)\widetilde k.
\end{equation}

Because $I-P$ is symmetric and positive semidefinite, Cauchy--Schwarz under its induced seminorm gives
\begin{equation}
    \left(
        \widetilde q^\top(I-P)\widetilde k
    \right)^2
    \leq
    \left(
        \widetilde q^\top(I-P)\widetilde q
    \right)
    \left(
        \widetilde k^\top(I-P)\widetilde k
    \right)
    =
    XY.
\end{equation}

Taking expectations and using the product-of-marginals factorization in Equation~\ref{eq:app_product_factorization},
\begin{equation}
    L_{\mathrm{sel}}(P)
    \leq
    \mathbb{E}[XY]
    =
    \mathbb{E}[X]\mathbb{E}[Y].
\end{equation}

Applying AM--GM yields
\begin{equation}
    \mathbb{E}[X]\mathbb{E}[Y]
    \leq
    \frac{1}{4}
    \left(
        \mathbb{E}[X]
        +
        \mathbb{E}[Y]
    \right)^2.
\end{equation}

Therefore,
\begin{equation}
    L_{\mathrm{sel}}(P)
    \leq
    \frac{1}{4}
    L_{\mathrm{marg}}(P)^2,
    \label{eq:app_marginal_bound}
\end{equation}
where
\begin{align}
    L_{\mathrm{marg}}(P)
    &=
    \mathbb{E}[X]
    +
    \mathbb{E}[Y]
    \nonumber\\
    &=
    \operatorname{tr}
    \left(
        (I-P)
        \left(
            \widetilde G_{\bar Q}
            +
            \widetilde G_K
        \right)
    \right).
    \label{eq:app_marginal_objective}
\end{align}

The marginal objective has the closed-form rank-$r$ solution
\begin{equation}
    M^\star_{\mathrm{marg}}
    =
    \operatorname{TopEig}_r
    \left(
        \widetilde G_{\bar Q}
        +
        \widetilde G_K
    \right)^\top.
    \label{eq:app_closed_form_basis}
\end{equation}

For
$P^\star_{\mathrm{marg}}
=
M_{\mathrm{marg}}^{\star\top}
M^\star_{\mathrm{marg}}$,
Equation~\ref{eq:app_marginal_bound} implies
\begin{equation}
    L_{\mathrm{sel}}
    \left(
        P^\star_{\mathrm{marg}}
    \right)
    \leq
    \frac{1}{4}
    \left[
        \min_{\substack{
            P=P^\top=P^2\\
            \operatorname{rank}(P)=r
        }}
        L_{\mathrm{marg}}(P)
    \right]^2.
\end{equation}

This bound controls the exact score error but does not imply that $P^\star_{\mathrm{marg}}$ globally minimizes $L_{\mathrm{sel}}$ or that the exact and marginal objectives share a minimizer.

The derivation does not require $\widetilde G_{\bar Q}$ and $\widetilde G_K$ to commute.
Their normalized commutator ratio on LLaMA-3.1-8B is $0.470$ on average and $0.476$ at the median, confirming that the shared-eigenbasis special case does not hold empirically.

\subsection{Exact-Objective Refinement}
\label{app:objective_refinement}

The marginal bound does not establish that the closed-form basis
$M^\star_{\mathrm{marg}}$ globally minimizes the exact selection objective.
We therefore test whether direct optimization can improve upon the spectral solution.

Starting from
\begin{equation}
    M^\star_{\mathrm{marg}}
    =
    \operatorname{TopEig}_r
    \left(
        \widetilde G_{\bar Q}
        +
        \widetilde G_K
    \right)^\top,
\end{equation}
we optimize $L_{\mathrm{sel}}$ for 1,600 steps using retraction-based gradient descent on the row-orthonormal Stiefel manifold, with no momentum.
For LLaMA-3.1-8B-Instruct, we optimize all 256 layer--KV-head-specific bases in parallel, with one independent objective and basis for each layer--KV-head pair.
The normalized second-moment matrices are computed from the same eight 128K-token LA-packed calibration sequences used throughout the paper.
At each step, we take a Euclidean gradient update and map the result back to the manifold using a QR retraction:
\begin{equation}
    M_{t+1}^{\top}
    =
    \operatorname{qf}
    \left(
        \left[
            M_t
            -
            \eta
            \nabla_M
            L_{\mathrm{sel}}(M_t)
        \right]^\top
    \right),
    \label{eq:app_qr_retraction}
\end{equation}
where $\operatorname{qf}(\cdot)$ denotes the orthonormal factor of a QR decomposition.
Because $L_{\mathrm{sel}}$ is a positive rescaling of the unnormalized score error (Equation~\ref{eq:app_normalized_exact_objective}), this refinement targets the same minimizers as the unnormalized objective; trace normalization affects only the spectral surrogate used for initialization.

\begin{table}[h]
\centering
\caption{
Exact-objective refinement from the spectral initialization on LLaMA-3.1-8B-Instruct.
We sweep the learning rate over three orders of magnitude and report the exact objective before and after 1,600 optimization steps.
}
\label{tab:exact_refinement_loss}
\vspace{4pt}

{\renewcommand{\arraystretch}{0.95}
\setlength{\tabcolsep}{5.5pt}

\begin{tabular}{ccrrrr}
\toprule
$r$
&
$\eta$
&
$L_{\mathrm{sel}}^{(0)}$
&
$L_{\mathrm{sel}}^{(1600)}$
&
Relative Change
&
Overlap
\\
\midrule
\multirow{4}{*}{8}
& 0.005 & $1.752351{\times}10^{-3}$ & $1.751730{\times}10^{-3}$ & $-0.04\%$ & 0.999999 \\
& 0.05  & $1.752351{\times}10^{-3}$ & $1.747971{\times}10^{-3}$ & $-0.25\%$ & 0.999958 \\
& 0.5   & $1.752351{\times}10^{-3}$ & $1.736314{\times}10^{-3}$ & $-0.92\%$ & 0.998841 \\
& 5.0   & $1.752351{\times}10^{-3}$ & $1.724382{\times}10^{-3}$ & $-1.60\%$ & 0.989120 \\
\midrule
\multirow{4}{*}{32}
& 0.005 & $3.968264{\times}10^{-4}$ & $3.967997{\times}10^{-4}$ & $-0.01\%$ & 1.000000 \\
& 0.05  & $3.968264{\times}10^{-4}$ & $3.965928{\times}10^{-4}$ & $-0.06\%$ & 0.999999 \\
& 0.5   & $3.968264{\times}10^{-4}$ & $3.955870{\times}10^{-4}$ & $-0.31\%$ & 0.999972 \\
& 5.0   & $3.968264{\times}10^{-4}$ & $3.935436{\times}10^{-4}$ & $-0.83\%$ & 0.999476 \\
\bottomrule
\end{tabular}
}
\end{table}

Table~\ref{tab:exact_refinement_loss} shows that the conclusion is insensitive to the step size.
Across a $1000\times$ learning-rate range, direct refinement reduces the exact objective by at most $1.60\%$ at $r=8$ and $0.83\%$ at $r=32$.
The refined subspace also remains close to the spectral initialization; at $r=32$, the overlap remains at least $0.999476$ throughout the sweep.

This behavior is not caused by a stalled optimization procedure.
Using the same Euclidean-gradient update and QR retraction with $\eta=0.05$, random initialization reduces the objective by $15.0\times$ at $r=8$ and $16.9\times$ at $r=32$ over 1,600 steps.
Nevertheless, its final objective remains $3.53\times$ and $8.97\times$ larger than that of the spectral initialization, respectively, and its overlap with the spectral subspace is only 0.242 and 0.287.
For diagnostic purposes, we additionally measure the norm of the tangent-projected gradient, which is not used directly in the update rule.
At initialization, this norm is $354\times$ larger for a random basis at $r=8$ and $2{,}412\times$ larger at $r=32$ than for the spectral basis.
These controls confirm that optimization is active, while the spectral solution already lies in a flat, low-objective region that is difficult to reach from a generic initialization within the same optimization budget.

\begin{table}[h]
\centering
\caption{
Selection stability under exact-objective refinement.
Starting from $M^\star_{\mathrm{marg}}$, we optimize the exact objective for 1,600 steps with $\eta=0.05$.
Oracle top-$B$ recall and subspace overlap remain essentially unchanged.
For Gemma4-26B-A4B-It, results are reported over the five global-attention layers to which JSSA is applied.
}
\label{tab:exact_refinement}
\vspace{4pt}

{\renewcommand{\arraystretch}{0.95}
\setlength{\tabcolsep}{8pt}

\scalebox{0.95}{
\begin{tabular}{lccccc}
\toprule
\multirow{2}{*}{Model}
&
\multirow{2}{*}{$r$}
&
\multirow{2}{*}{$B$}
&
\multicolumn{2}{c}{Oracle Recall}
&
\multirow{2}{*}{Subspace Overlap}
\\
\cmidrule(lr){4-5}
&
&
&
Step 0
&
Step 1600
&
\\
\midrule

\multirow{4}{*}{LLaMA-3.1-8B-Instruct}
&
\multirow{2}{*}{8}
&
256
&
48.76\%
&
48.74\%
&
\multirow{2}{*}{0.999958}
\\

&
&
2048
&
53.81\%
&
53.81\%
&
\\

\cmidrule(lr){2-6}

&
\multirow{2}{*}{32}
&
256
&
71.34\%
&
71.36\%
&
\multirow{2}{*}{0.999999}
\\

&
&
2048
&
75.94\%
&
75.95\%
&
\\

\midrule

\multirow{2}{*}{Gemma4-26B-A4B-It}
&
\multirow{2}{*}{64}
&
256
&
79.85\%
&
79.85\%
&
\multirow{2}{*}{1.000000}
\\

&
&
2048
&
83.33\%
&
83.33\%
&
\\

\bottomrule
\end{tabular}
}
}
\end{table}

Table~\ref{tab:exact_refinement} reports oracle top-$B$ recall before and after refinement across two models, multiple projection ranks, and token budgets $B\in\{256,2048\}$.
For Gemma4-26B-A4B-It, refinement is evaluated at the default rank $r=64$ over the five global-attention layers to which JSSA is applied.
The change in recall is at most $0.02$ percentage points.

We measure the overlap between the initial and refined subspaces as
\begin{equation}
    \operatorname{Overlap}
    \left(
        M,M_{\mathrm{ref}}
    \right)
    =
    \frac{1}{r}
    \operatorname{tr}
    \left(
        PP_{\mathrm{ref}}
    \right),
    \qquad
    P=M^\top M,
    \quad
    P_{\mathrm{ref}}=M_{\mathrm{ref}}^\top M_{\mathrm{ref}}.
    \label{eq:app_subspace_overlap}
\end{equation}
This quantity is the mean squared cosine of the principal angles between the two subspaces and equals one when they are identical.

Together, these results show that direct optimization can marginally reduce the exact objective but does not measurably improve token-selection quality in the tested configurations.
They support using $M^\star_{\mathrm{marg}}$ directly in practice, while not establishing either the tightness of the marginal bound or the global optimality of the spectral solution.

\subsection{Oracle Selection Recall}
\label{app:oracle_selection_recall}

We evaluate how well each selected token set matches the oracle top-$B$ set obtained from full-dimensional attention scores.

For a method producing a selected set $\mathcal I_B$ and the oracle set $\mathcal I_B^\star$, we define oracle top-$B$ recall as
\begin{equation}
    \operatorname{Recall@}B
    =
    \frac{
        \left|
            \mathcal I_B
            \cap
            \mathcal I_B^\star
        \right|
    }{
        B
    }.
    \label{eq:app_oracle_recall}
\end{equation}

Table~\ref{tab:oracle_recall} compares JSSA with Loki and FASA on LLaMA-3.1-8B-Instruct and Gemma4-26B-A4B-It at their default projection ranks.
Each method uses the same calibration configuration as in the main evaluation: packed 128K LongAlign sequences for JSSA and Loki, and plain LongAlign examples for FASA.
JSSA achieves the highest oracle recall for both models and token budgets, outperforming Loki by 6.25--35.37 percentage points and FASA by 22.44--35.25 points.

These results complement the downstream evaluations in Section~\ref{sec:experiments}, showing that the joint query--key basis more faithfully preserves the token ranking induced by full-dimensional attention scores across model architectures and projection ranks.

\begin{table}[h]
\centering
\caption{
Oracle top-$B$ recall on LLaMA-3.1-8B-Instruct and Gemma4-26B-A4B-It at their default projection ranks.
}
\label{tab:oracle_recall}
\vspace{4pt}

{\renewcommand{\arraystretch}{0.95}
\setlength{\tabcolsep}{8pt}
\begin{tabular}{llccc}
\toprule
Model & $B$ & JSSA & Loki & FASA \\
\midrule

\multirow{2}{*}{LLaMA-3.1-8B-Instruct ($r=32$)}
& 256  & 71.34\% & 35.97\% & 48.90\% \\
& 2048 & 75.94\% & 51.84\% & 53.31\% \\
\midrule

\multirow{2}{*}{Gemma4-26B-A4B-It ($r=64$)}
& 256  & 79.85\% & 67.79\% & 44.60\% \\
& 2048 & 83.33\% & 77.08\% & 49.90\% \\
\bottomrule
\end{tabular}
}
\end{table}

\subsection{Comparison with Asymmetric Query--Key Factorization}
\label{app:kqsvd_comparison}

KQ-SVD~\citep{kqsvd} was originally proposed for KV-cache compression and derives separate low-dimensional query and key transformations to preserve attention scores.
To isolate whether this asymmetric score-preserving factorization is also effective for dynamic token selection, we adapt its factorization to JSSA's scoring pipeline rather than using KQ-SVD as a compression method.
Specifically, we compute the KQ-SVD-style query and key projections from the same calibration activations used by JSSA and use them only to score candidate tokens; the original full-dimensional KV cache, token budgets, selection granularity, and sparse-attention execution remain unchanged.
The controlled comparison therefore primarily isolates JSSA's shared orthonormal scoring subspace from an asymmetric query--key scoring factorization.

\paragraph{RULER.}
Table~\ref{tab:kqsvd_ruler} compares the two scoring representations on the same RULER configurations used in the main evaluation.
The asymmetric variant remains competitive in several settings, including closely matching JSSA on LLaMA-3.1-8B-Instruct and Ministral-3-8B-Instruct-2512.
However, its performance is less consistent across architectures and context lengths, with a particularly large gap on Qwen3-14B.
These results show that the asymmetric factorization provides a viable low-dimensional scoring representation rather than a uniformly weak baseline.

\begin{table}[h]
\centering
\caption{
RULER accuracy of JSSA and an asymmetric KQ-SVD factorization adapted to token scoring under the same fixed-density KV budgets as the main evaluation.
}
\label{tab:kqsvd_ruler}
\vspace{4pt}

{\renewcommand{\arraystretch}{0.95}
\setlength{\tabcolsep}{7pt}
\begin{tabular}{llccc}
\toprule
Model & Method & 32K & 64K & 128K \\
\midrule
\multirow{2}{*}{LLaMA-3.1-8B-Instruct}
& Asymmetric (KQ-SVD) & 84.25 & 81.77 & 73.23 \\
& JSSA   & \textbf{86.35} & \textbf{82.55} & \textbf{75.38} \\
\midrule
\multirow{2}{*}{Qwen3-14B}
& Asymmetric (KQ-SVD) & 71.46 & 82.42 & 75.78 \\
& JSSA   & \textbf{88.86} & \textbf{88.47} & \textbf{82.21} \\
\midrule
\multirow{2}{*}{Ministral-3-8B-Instruct-2512}
& Asymmetric (KQ-SVD) & 89.71 & 86.77 & \textbf{86.28} \\
& JSSA   & \textbf{90.72} & \textbf{89.02} & 85.97 \\
\bottomrule
\end{tabular}
}
\end{table}

\paragraph{Reasoning.}
The difference becomes more pronounced under aggressive sparsity on reasoning benchmarks.
Table~\ref{tab:kqsvd_reasoning} compares JSSA with the asymmetric KQ-SVD factorization adapted to token scoring on AIME 2025 and GPQA-Diamond for Qwen3-8B and Gemma4-26B-A4B-It.
At a budget of 512 tokens, the asymmetric variant obtains only 0.83\% and 26.67\% on AIME 2025, compared with 64.58\% and 88.75\% for JSSA.
The same trend appears on GPQA-Diamond, where it trails JSSA by 14.14 and 24.87 percentage points at 512 tokens.
Its accuracy recovers substantially as the token budget increases and becomes competitive in several 2,048-token settings.
These results indicate that the asymmetric factorization can provide useful token scores, but is substantially less robust than JSSA's shared orthonormal subspace under aggressive sparsity.

\begin{table}[h]
\centering
\caption{
Reasoning pass@1 accuracy of JSSA and an asymmetric KQ-SVD factorization adapted to token scoring across token budgets.
}
\label{tab:kqsvd_reasoning}
\vspace{4pt}

{\renewcommand{\arraystretch}{0.95}
\setlength{\tabcolsep}{5pt}
\begin{tabular}{llcccccc}
\toprule
& & \multicolumn{3}{c}{AIME 2025} & \multicolumn{3}{c}{GPQA-Diamond} \\
Model & Method & 512 & 1024 & 2048 & 512 & 1024 & 2048 \\
\midrule
\multirow{2}{*}{Qwen3-8B}
& Asymmetric (KQ-SVD) & 0.83 & 9.17 & 34.58 & 43.94 & 53.54 & \textbf{59.09} \\
& JSSA   & \textbf{64.58} & \textbf{68.33} & \textbf{67.92} & \textbf{58.08} & \textbf{57.39} & 58.21 \\
\midrule
\multirow{2}{*}{Gemma4-26B-A4B-It}
& Asymmetric (KQ-SVD) & 26.67 & 77.50 & 85.83 & 55.81 & 75.25 & \textbf{81.82} \\
& JSSA   & \textbf{88.75} & \textbf{86.67} & \textbf{88.33} & \textbf{80.68} & \textbf{81.94} & 81.12 \\
\bottomrule
\end{tabular}
}
\end{table}

Overall, these results isolate a substantial robustness difference between shared and asymmetric query--key scoring parameterizations.
The asymmetric variant remains competitive in several RULER settings and recovers at larger reasoning budgets, but degrades sharply under aggressive sparsity on reasoning benchmarks.
One possible explanation is that the more expressive asymmetric factorization overfits calibration-specific query--key correlations, whereas JSSA's shared orthonormal subspace provides stronger regularization across query--key combinations.
This behavior supports JSSA's design of constructing a single scoring subspace that jointly reflects the query and key distributions.
We leave a direct theoretical and empirical characterization of this robustness difference to future work.

\newpage

\section{Large-Scale Validation}
\label{app:large_scale}

Section~\ref{sec:large_scale} reports headline accuracy and throughput results comparing post-hoc JSSA with model-native DSA at approximately 700B scale. Here we provide the complete experimental setup, additional context, and full throughput sweeps for both model families. Unless otherwise stated, these experiments use eight NVIDIA B200 GPUs with tensor parallelism degree eight.

\paragraph{DeepSeek.}
We compare two checkpoints from the same model lineage.
DeepSeek-V3.1-Terminus uses full attention, whereas DeepSeek-V3.2-Exp introduces native DSA through continued pre-training.
We evaluate DeepSeek-V3.1-Terminus with FKV attention and with JSSA applied post hoc, and compare them with the officially released DeepSeek-V3.2-Exp checkpoint using native DSA.
DeepSeek-V3.2-Exp is derived from DeepSeek-V3.1-Terminus with DSA as its only architectural modification and continued training under aligned configurations.
This comparison therefore follows DeepSeek's intended design for evaluating native sparse attention, with native DSA additionally adapted to its sparse attention pattern through continued training.
JSSA and native DSA use scoring representations with the same total dimensionality of 128, but with different constructions.
JSSA uses 64 latent and 64 RoPE dimensions, whereas native DSA uses learned 128-dimensional lightning-indexer heads.
Both methods select 2,048 tokens at every attention layer.

Table~\ref{tab:large_scale_reasoning} in Section~\ref{sec:large_scale} reports pass@1 accuracy on AIME 2025 and GPQA-Diamond.
Given the small numerical differences involved and the potential for evaluation variance, we interpret these results as showing comparable reasoning accuracy rather than a consistent advantage for any individual method.

\begin{figure}[t]
\centering
\vspace{-4pt}
\includegraphics[width=0.95\textwidth]{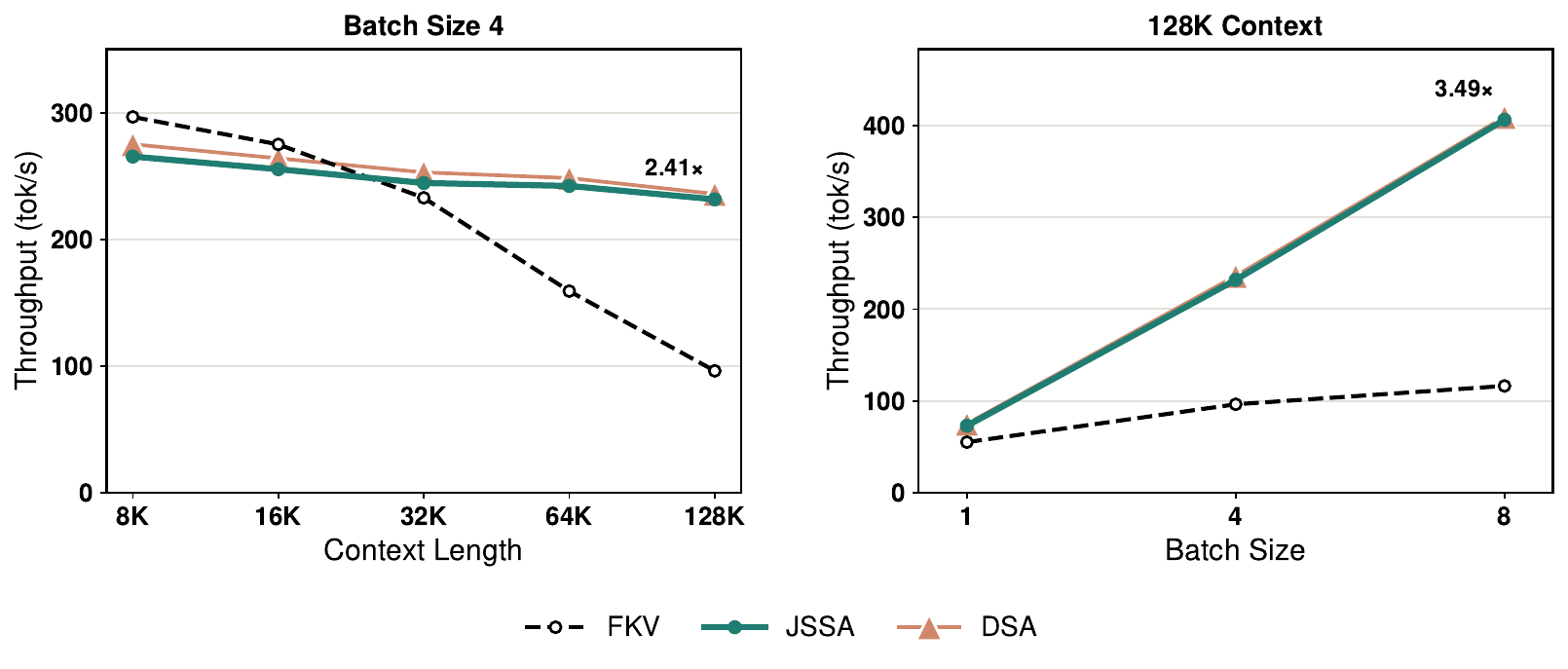}
\vspace{-12pt}
\caption{
\textbf{End-to-end decoding throughput on DeepSeek.}
Throughput across context lengths at batch size 4 (left) and across batch sizes at 128K context (right), measured on eight NVIDIA B200 GPUs with tensor parallelism degree eight.
}
\label{fig:deepseek_throughput}
\end{figure}

Figure~\ref{fig:deepseek_throughput} reports end-to-end decoding throughput.
Across the complete sweep, JSSA reaches 96.1--99.2\% of native DSA throughput.
At batch size 4, its throughput relative to FKV increases from $0.89\times$ at 8K context to $2.41\times$ at 128K.
At 128K context, the speedup reaches $3.49\times$ at batch size 8.

\paragraph{GLM-5.2.}
We additionally evaluate JSSA on GLM-5.2, an independently developed MLA model with native DSA.
GLM-5.2 invokes its DSA indexer only at a subset of attention layers and reuses each selected token set in adjacent layers to reduce indexing overhead.
JSSA follows the same native schedule, performing selection only where the DSA indexer is invoked and sharing the resulting indices with the corresponding adjacent layers.
Because no corresponding full-attention checkpoint is available, we compare only JSSA and native DSA.

As shown in Table~\ref{tab:large_scale_reasoning}, their AIME 2025 scores differ by 0.83 percentage points, while their GPQA-Diamond scores are identical, providing further evidence that post-hoc JSSA matches native DSA's reasoning accuracy under a model-native sparse-attention schedule that reuses selected indices across layers.

\section{Decoding Throughput Details}
\label{app:throughput_details}

\begin{figure}[h]
\centering
\vspace{-4pt}
\includegraphics[width=1.0\textwidth]{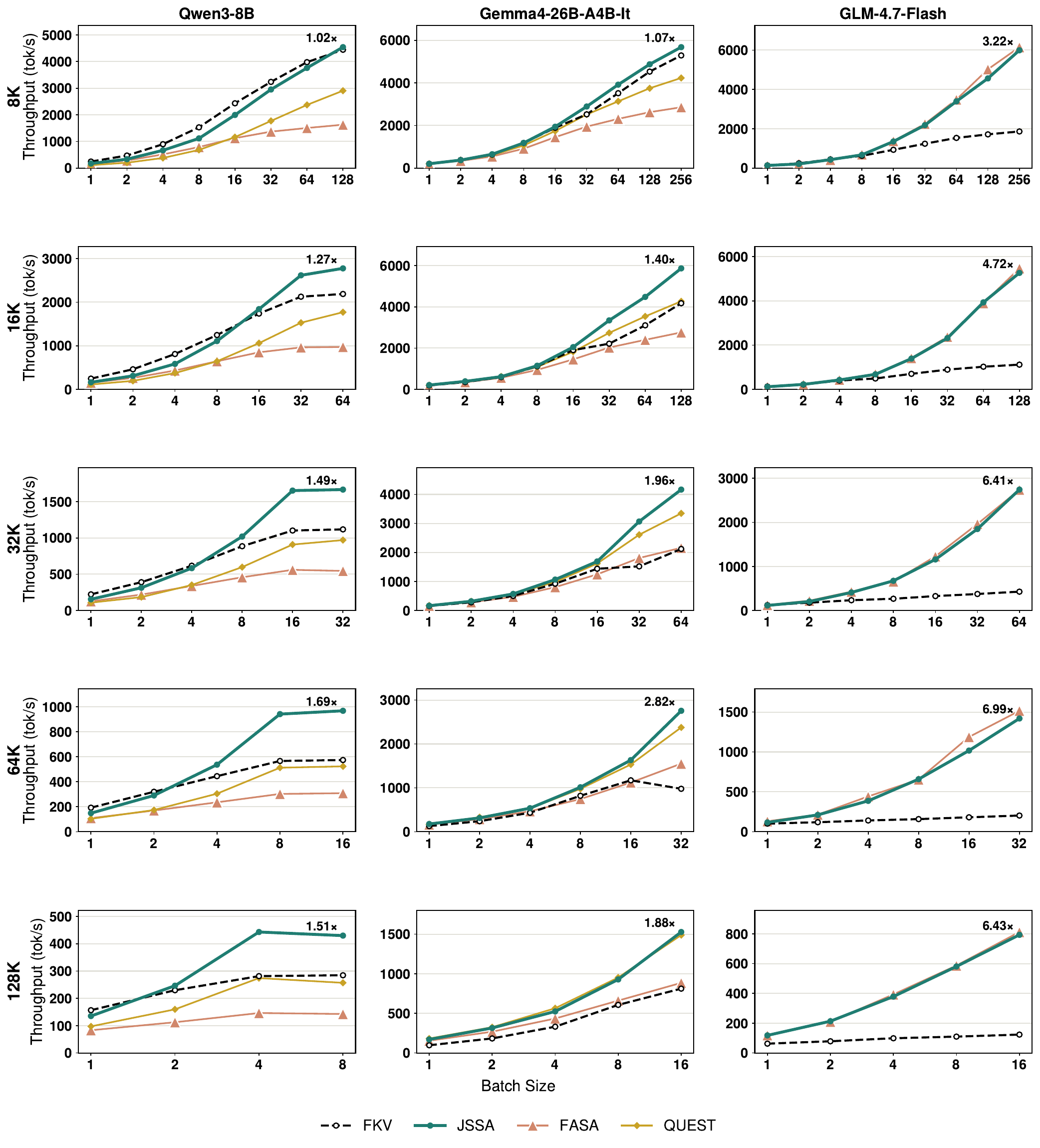}
\vspace{-16pt}
\caption{
\textbf{Complete end-to-end decoding-throughput sweeps.}
Results across context lengths and batch sizes on a single NVIDIA B200 GPU.
Annotations show JSSA's maximum speedup over FKV in each panel.
}
\label{fig:throughput_full}
\end{figure}

Figure~\ref{fig:throughput_full} presents the complete end-to-end decoding-throughput sweeps across context lengths and batch sizes for Qwen3-8B, Gemma4-26B-A4B-It, and GLM-4.7-Flash.
We follow the measurement protocol in Appendix~\ref{app:hardware_protocol}.

Across all three models, JSSA's benefit generally increases with context length and batch size as KV-cache traffic becomes a larger fraction of decoding time.
At shorter contexts and smaller batches, scoring and top-$B$ selection can offset part of the reduction in KV access, particularly on Qwen3-8B.
As the workload becomes more memory-bandwidth intensive, JSSA consistently surpasses full-KV attention and achieves larger gains.

The magnitude of the speedup depends on the attention architecture.
Across the complete sweep, JSSA reaches up to $1.69\times$, $2.82\times$, and $6.99\times$ the throughput of full-KV attention on Qwen3-8B, Gemma4-26B-A4B-It, and GLM-4.7-Flash, respectively.
The Gemma4 result is obtained while sparsifying only the global-attention layers, whereas the larger GLM-4.7-Flash gain benefits from sparse FlashMLA execution over its shared latent cache.
These results complement the representative 32K and 128K settings in Figure~\ref{fig:throughput} and show that the gains persist across a broad range of serving configurations.

\newpage

\section{Additional Experimental Results}
\label{app:additional_results}

\subsection{Component-Level Latency Breakdown}
\label{app:component_breakdown}

\begin{figure}[h]
\centering
\vspace{-8pt}
\includegraphics[width=1.0\textwidth]{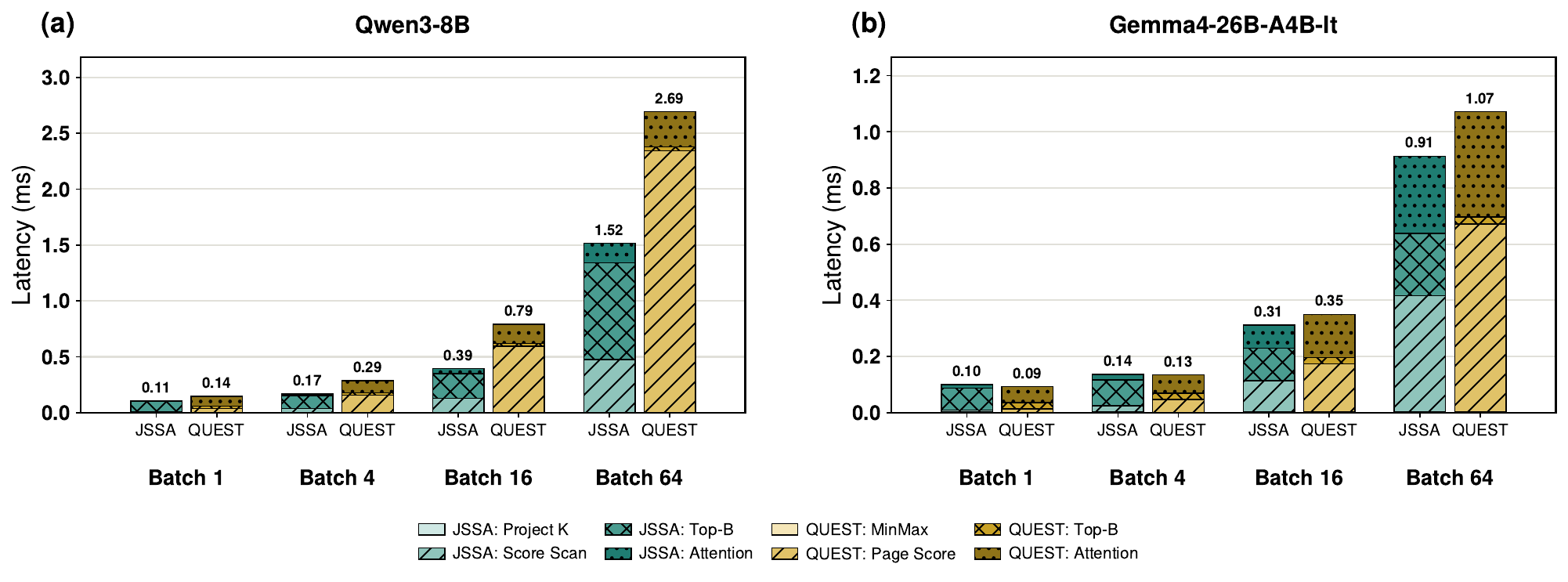}
\vspace{-18pt}
\caption{Component-level sparse-attention latency at 128K context.
Bars decompose JSSA and QUEST into scoring preparation, candidate scoring, top-$B$ selection, and sparse attention.
JSSA's key-projection overhead is negligible, and its latency advantage over QUEST increases at larger batches.}
\label{fig:component_breakdown}
\end{figure}

Figure~\ref{fig:component_breakdown} decomposes the sparse-attention pipeline at 128K context into scoring preparation, candidate scoring, top-$B$ selection, and sparse attention.
For JSSA, projecting the newly generated key is negligible, accounting for at most 3.2\% of pipeline latency and less than 1\% at larger batches.
Top-$B$ selection dominates at small batches, while the full-context score scan becomes increasingly important as the batch grows.

On Qwen3-8B, JSSA is faster than QUEST at every measured batch size, reducing pipeline latency from 2.69\,ms to 1.52\,ms at batch size 64.
On Gemma4-26B-A4B-It, the two methods are comparable at small batches, but JSSA becomes faster as the batch increases, reaching 0.91\,ms versus 1.07\,ms for QUEST at batch size 64.
QUEST's page-scoring cost grows more rapidly with batch size, whereas JSSA benefits from its compact projected cache and regular score-scan kernel.

\subsection{Projected Scoring Cache and Quantization}
\label{app:scoring_cache}

JSSA stores the auxiliary projected-key cache used for token scoring in FP8, while the original full-dimensional keys and values used by final attention remain in BF16.
To isolate the effect of scoring-cache quantization, we compare FP8 and BF16 projected-key caches on LLaMA-3.1-8B-Instruct, Qwen3-14B, and Ministral-3-8B-Instruct-2512 at the default projection rank $r=32$.
All other configurations, including the projection basis, token budget, and final attention computation, are held fixed.

\begin{figure}[h]
\centering
\vspace{-8pt}
\includegraphics[width=0.5\textwidth]{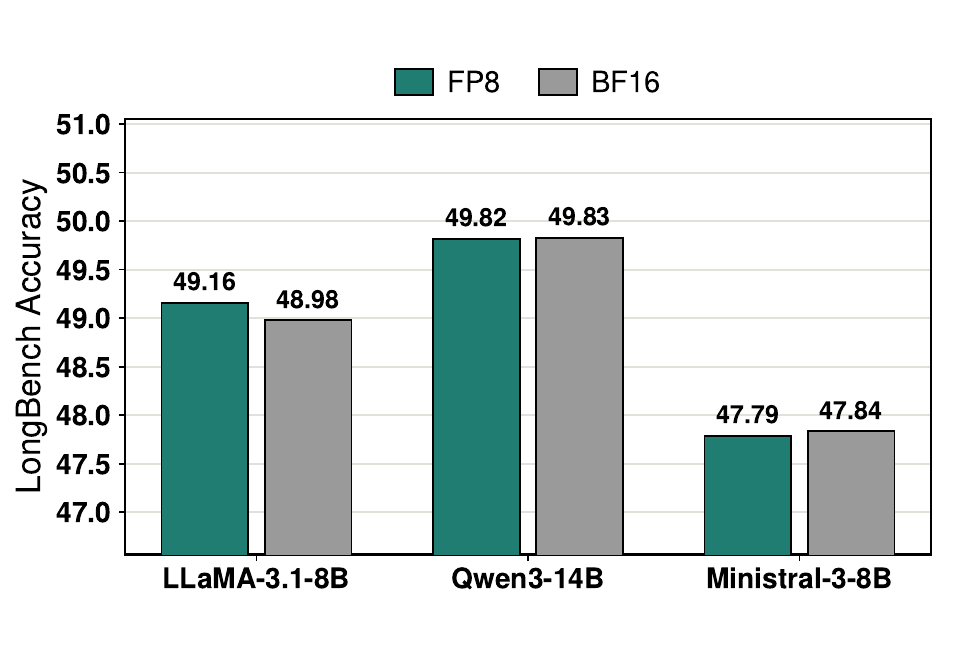}
\vspace{-18pt}
\caption{
\textbf{Effect of projected scoring-cache precision.}
Average LongBench accuracy at a KV budget of 256 tokens using FP8 and BF16 projected-key caches.
The original full-dimensional KV cache and final attention computation remain in BF16.
}
\label{fig:scoring_cache_quantization}
\end{figure}

Figure~\ref{fig:scoring_cache_quantization} reports average LongBench accuracy at a KV budget of 256 tokens.
FP8 and BF16 yield nearly identical results on all three models: the absolute difference is 0.18 points on LLaMA-3.1-8B-Instruct, 0.01 points on Qwen3-14B, and 0.05 points on Ministral-3-8B-Instruct-2512.
These results show that storing the projected scoring cache in FP8 has negligible impact on accuracy while reducing its memory footprint by half relative to BF16.

\subsection{Selection Granularity in GQA}
\label{app:gqa_selection_granularity}

In GQA, JSSA selects a distinct token set for each KV head using the mean of its associated query heads.
We compare this design with a head-shared alternative inspired by TokenSelect~\cite{tokenselect}, which aggregates head-specific token preferences through soft voting before forming a single token set for the entire layer.
This is stronger than naive unweighted sharing, but it still collapses distinct KV-head preferences into one shared selection.

Table~\ref{tab:gqa_selection_granularity} reports RULER accuracy under the same fixed-density budgets used in the main evaluation.
At 32K context, head-shared soft voting performs nearly identically to per-KV-head selection on all three models.
At 128K, however, per-KV-head selection is consistently better, by 5.29 points on LLaMA-3.1-8B-Instruct, 0.99 points on Ministral-3-8B-Instruct-2512, and 1.32 points on Qwen3-14B.

These results show that preserving distinct per-KV-head selections becomes important at long context.
To our knowledge, existing gather-free token-level kernels do not support such per-KV-head index sets under the standard GQA KV-cache layout.
JSSA's indexed-attention kernel enables this more accurate selection granularity without physically gathering the selected keys and values.

\begin{table}[t]
\centering
\caption{
RULER accuracy of per-KV-head selection and head-shared soft voting in GQA.
The budgets are 512, 1,024, and 2,048 tokens at 32K, 64K, and 128K context, respectively.
$\Delta$ denotes head-shared minus per-KV-head accuracy.
}
\label{tab:gqa_selection_granularity}
\vspace{4pt}

{\renewcommand{\arraystretch}{1.0}
\setlength{\tabcolsep}{6pt}
\begin{tabular}{llrrr}
\toprule
Model & Selection & 32K & 64K & 128K \\
\midrule

\multirow{3}{*}{LLaMA-3.1-8B-Instruct}
& Per-KV-head & 86.35 & 82.55 & 75.38 \\
& Head-shared soft vote & 86.03 & 80.82 & 70.09 \\
& $\Delta$ & -0.32 & -1.73 & -5.29 \\
\midrule

\multirow{3}{*}{Ministral-3-8B-Instruct-2512}
& Per-KV-head & 90.72 & 89.02 & 85.97 \\
& Head-shared soft vote & 90.67 & 87.24 & 84.98 \\
& $\Delta$ & -0.05 & -1.78 & -0.99 \\
\midrule

\multirow{3}{*}{Qwen3-14B}
& Per-KV-head & 88.86 & 88.47 & 82.21 \\
& Head-shared soft vote & 89.11 & 88.82 & 80.89 \\
& $\Delta$ & +0.25 & +0.35 & -1.32 \\
\bottomrule
\end{tabular}
}
\end{table}

\newpage

\subsection{Complete Long-Context Results}
\label{app:complete_long_context}

\subsubsection{LongBench}
\label{app:longbench_full}

Table~\ref{tab:longbench_full} reports the complete task-level LongBench results underlying the model-level averages in Table~\ref{tab:longbench}.
All sparse methods use a KV budget of 256 tokens, and FASA, Loki, and JSSA use their default 32-dimensional scoring configurations.
JSSA is consistently competitive across task categories and achieves the best sparse average on LLaMA-3.1-8B and Qwen3-14B, while remaining close to the strongest sparse baselines on Ministral-3-8B.
The detailed results also show that competing methods tend to exhibit larger task- and model-specific variation, whereas JSSA remains comparatively stable across single-document QA, multi-document QA, summarization, few-shot learning, synthetic tasks, and code.

\begin{table}[h]
\centering
\caption{Complete task-level LongBench results at a KV budget of 256 tokens.
The final column reports the average across all tasks.}
\label{tab:longbench_full}
\tiny
\setlength{\tabcolsep}{1.25pt}
\renewcommand{\arraystretch}{1.05}
\newcommand{\rot}[1]{\rotatebox{75}{\scriptsize #1}}
\resizebox{\linewidth}{!}{
\begin{tabular}{lccccccccccccccccc}
\toprule
& \multicolumn{3}{c}{Single-Document QA}
& \multicolumn{3}{c}{Multi-Document QA}
& \multicolumn{3}{c}{Summarization}
& \multicolumn{3}{c}{Few-Shot Learning}
& \multicolumn{2}{c}{Synthetic}
& \multicolumn{2}{c}{Code}
& \multirow{2}{*}{Avg.} \\
\cmidrule(lr){2-4}
\cmidrule(lr){5-7}
\cmidrule(lr){8-10}
\cmidrule(lr){11-13}
\cmidrule(lr){14-15}
\cmidrule(lr){16-17}
Method
& \rot{NarrativeQA}
& \rot{Qasper}
& \rot{MultiFieldQA-en}
& \rot{HotpotQA}
& \rot{2WikiMQA}
& \rot{Musique}
& \rot{GovReport}
& \rot{QMSum}
& \rot{MultiNews}
& \rot{TREC}
& \rot{TriviaQA}
& \rot{SAMSum}
& \rot{PassageCount}
& \rot{PassageRetrieval-en}
& \rot{LCC}
& \rot{RepoBench-P}
& \\
\midrule

\multicolumn{18}{c}{\textit{LLaMA-3.1-8B-Instruct}} \\
\midrule
FKV
& 29.67 & 45.58 & 55.09
& 56.47 & 45.07 & 32.23
& 35.32 & 24.94 & 27.17
& 72.50 & 91.48 & 43.58
& 7.93 & 99.50
& 54.72 & 52.28
& 48.35 \\
SnapKV
& 25.70 & 31.04 & 51.46
& 54.50 & 42.70 & 26.86
& 22.87 & 23.60 & 22.53
& 55.50 & 91.57 & 41.58
& 8.42 & 99.50
& 53.85 & 44.84
& 43.53 \\
QUEST
& 29.21 & 43.63 & 54.39
& 54.89 & 44.48 & 28.53
& 32.66 & 24.51 & 26.41
& 72.00 & 91.67 & 42.21
& 8.04 & 98.00
& 54.00 & 46.40
& 46.94 \\
SqueezedAttn.
& 29.32 & 45.20 & 55.31
& 55.43 & 46.02 & 30.45
& 35.41 & 24.30 & 27.14
& 72.50 & 91.75 & 42.03
& 8.65 & 99.00
& 54.64 & 50.12
& 47.95 \\
FASA
& 26.31 & 43.62 & 53.65
& 52.87 & 42.12 & 29.33
& 31.59 & 25.05 & 27.63
& 70.00 & 89.64 & 44.26
& 7.70 & 99.50
& 55.98 & 48.57
& 46.74 \\
Loki
& 25.91 & 39.34 & 52.85
& 52.73 & 41.06 & 28.65
& 29.00 & 22.96 & 27.19
& 72.50 & 89.85 & 43.08
& 7.78 & 99.00
& 56.00 & 42.49
& 45.65 \\
\rowcolor{teal!10}
JSSA
& 30.32 & 44.72 & 55.20
& 55.69 & 45.51 & 30.68
& 35.05 & 24.86 & 26.89
& 72.50 & 91.27 & 43.77
& 7.31 & 99.50
& 66.28 & 56.99
& 49.16 \\

\midrule
\multicolumn{18}{c}{\textit{Qwen3-14B}} \\
\midrule
FKV
& 30.12 & 43.92 & 51.35
& 63.44 & 53.39 & 41.92
& 33.61 & 23.96 & 24.63
& 77.50 & 89.51 & 42.00
& 1.75 & 99.50
& 68.22 & 63.05
& 50.49 \\
SnapKV
& 28.52 & 36.76 & 47.51
& 62.51 & 50.33 & 37.31
& 24.18 & 22.54 & 20.57
& 59.50 & 88.91 & 38.97
& 2.40 & 99.50
& 56.11 & 48.71
& 45.27 \\
QUEST
& 30.00 & 41.31 & 51.29
& 59.76 & 53.03 & 39.29
& 31.88 & 22.83 & 24.61
& 71.50 & 89.55 & 40.71
& 1.60 & 98.50
& 65.48 & 53.73
& 48.44 \\
SqueezedAttn.
& 28.17 & 41.88 & 50.27
& 63.12 & 53.60 & 41.35
& 32.91 & 23.59 & 24.51
& 77.00 & 90.13 & 40.59
& 1.88 & 99.50
& 56.79 & 51.99
& 48.58 \\
FASA
& 26.36 & 41.91 & 50.30
& 56.96 & 51.47 & 38.34
& 31.76 & 23.86 & 24.84
& 71.50 & 87.56 & 40.81
& 1.28 & 97.25
& 65.86 & 53.88
& 47.93 \\
Loki
& 29.15 & 41.56 & 51.03
& 63.23 & 52.64 & 39.41
& 33.34 & 24.01 & 25.06
& 76.50 & 90.48 & 41.59
& 4.48 & 99.50
& 63.62 & 56.96
& 49.54 \\
\rowcolor{teal!10}
JSSA
& 28.87 & 41.68 & 50.95
& 64.03 & 52.81 & 40.60
& 32.84 & 23.47 & 24.64
& 77.00 & 90.40 & 41.80
& 1.70 & 99.50
& 66.38 & 60.43
& 49.82 \\

\midrule
\multicolumn{18}{c}{\textit{Ministral-3-8B-Instruct-2512}} \\
\midrule
FKV
& 26.69 & 43.72 & 44.50
& 57.79 & 49.47 & 36.05
& 22.51 & 15.94 & 19.72
& 75.50 & 89.81 & 45.04
& 12.50 & 98.50
& 68.53 & 60.20
& 47.90 \\
SnapKV
& 26.36 & 33.21 & 40.81
& 57.14 & 51.54 & 39.18
& 18.80 & 16.52 & 15.67
& 64.00 & 91.03 & 45.22
& 9.00 & 99.08
& 55.05 & 54.45
& 44.82 \\
QUEST
& 26.75 & 42.55 & 40.65
& 56.81 & 49.44 & 30.88
& 22.72 & 15.74 & 19.45
& 74.50 & 90.14 & 41.41
& 12.50 & 98.17
& 64.06 & 57.89
& 46.48 \\
SqueezedAttn.
& 27.37 & 44.89 & 45.60
& 61.09 & 54.23 & 38.89
& 22.72 & 17.56 & 19.20
& 77.00 & 90.79 & 45.58
& 9.50 & 99.50
& 56.92 & 55.71
& 47.91 \\
FASA
& 26.49 & 40.40 & 43.44
& 54.31 & 51.15 & 33.43
& 29.83 & 18.78 & 22.54
& 72.50 & 89.79 & 45.41
& 9.00 & 92.83
& 54.02 & 52.73
& 46.04 \\
Loki
& 28.68 & 42.17 & 45.83
& 58.11 & 53.76 & 37.27
& 22.27 & 16.94 & 19.91
& 74.50 & 91.46 & 42.77
& 7.50 & 100.00
& 66.65 & 61.36
& 48.07 \\
\rowcolor{teal!10}
JSSA
& 28.91 & 42.77 & 46.05
& 57.11 & 53.07 & 36.73
& 22.19 & 16.65 & 19.80
& 74.50 & 91.46 & 43.23
& 6.00 & 100.00
& 65.71 & 60.38
& 47.79 \\
\bottomrule
\end{tabular}
}
\end{table}

\subsubsection{RULER}
\label{app:ruler_full}

Table~\ref{tab:ruler_full} reports the complete task-level RULER results corresponding to Figure~\ref{fig:ruler}.
We use a fixed 1.5625\% KV budget, selecting 512, 1,024, and 2,048 tokens at context lengths of 32K, 64K, and 128K, respectively.
JSSA remains close to full-KV attention across models and context lengths, while several competing methods degrade sharply on particular retrieval and aggregation tasks.
The task-level breakdown shows that JSSA consistently preserves the needle-in-a-haystack variants, including the more challenging multi-key and multi-value settings, and avoids the stronger long-context degradation observed for Loki and FASA on LLaMA-3.1-8B.
On Ministral-3-8B, JSSA is within 2 points of full-KV average accuracy at 32K and 64K, and matches it at 128K.

\begin{table}[h]
\centering
\caption{Complete task-level RULER accuracy under a fixed 1.5625\% KV budget.
The \textit{Avg.} column reports the average across the 13 RULER tasks.}
\label{tab:ruler_full}
\tiny
\setlength{\tabcolsep}{5pt}
\renewcommand{\arraystretch}{0.93}
\newcommand{\rulerrot}[1]{\rotatebox{75}{\scriptsize #1}}
\scalebox{0.95}{
\begin{tabular}{lcccccccccccccc}
\toprule
Method
& Avg.
& \rulerrot{NIAH-S1}
& \rulerrot{NIAH-S2}
& \rulerrot{NIAH-S3}
& \rulerrot{NIAH-MK1}
& \rulerrot{NIAH-MK2}
& \rulerrot{NIAH-MK3}
& \rulerrot{NIAH-MV}
& \rulerrot{NIAH-MQ}
& \rulerrot{VT}
& \rulerrot{CWE}
& \rulerrot{FWE}
& \rulerrot{QA1}
& \rulerrot{QA2} \\
\midrule

\multicolumn{15}{c}{\textit{LLaMA-3.1-8B-Instruct}} \\
\midrule
\multicolumn{15}{c}{32K Context} \\
\midrule
FKV
& 87.95 & 100.00 & 100.00 & 100.00 & 100.00 & 96.00 & 100.00 & 99.00 & 100.00 & 95.20 & 45.20 & 92.00 & 72.00 & 44.00 \\
SnapKV
& 65.05 & 100.00 & 100.00 & 4.00 & 100.00 & 88.00 & 0.00 & 85.00 & 92.00 & 90.40 & 3.60 & 62.67 & 76.00 & 44.00 \\
QUEST
& 74.92 & 100.00 & 100.00 & 100.00 & 100.00 & 80.00 & 4.00 & 95.00 & 99.00 & 94.40 & 6.80 & 78.70 & 68.00 & 48.00 \\
SqueezedAttn.
& 83.14 & 100.00 & 100.00 & 100.00 & 100.00 & 96.00 & 56.00 & 99.00 & 100.00 & 93.60 & 61.60 & 50.67 & 76.00 & 48.00 \\
FASA
& 82.77 & 100.00 & 100.00 & 92.00 & 100.00 & 96.00 & 64.00 & 97.00 & 100.00 & 90.40 & 54.00 & 62.67 & 72.00 & 48.00 \\
Loki
& 79.27 & 100.00 & 100.00 & 76.00 & 96.00 & 92.00 & 36.00 & 94.00 & 99.00 & 79.20 & 51.60 & 74.70 & 80.00 & 52.00 \\
\rowcolor{teal!10}
JSSA
& 86.35 & 100.00 & 100.00 & 100.00 & 100.00 & 96.00 & 96.00 & 94.00 & 100.00 & 93.60 & 57.60 & 65.30 & 72.00 & 48.00 \\
\midrule
\multicolumn{15}{c}{64K Context} \\
\midrule
FKV
& 85.56 & 100.00 & 100.00 & 100.00 & 100.00 & 100.00 & 100.00 & 98.00 & 100.00 & 92.00 & 15.60 & 86.70 & 68.00 & 52.00 \\
SnapKV
& 61.56 & 100.00 & 92.00 & 16.00 & 100.00 & 76.00 & 0.00 & 62.00 & 90.00 & 87.20 & 2.40 & 54.67 & 68.00 & 52.00 \\
QUEST
& 76.35 & 100.00 & 96.00 & 100.00 & 100.00 & 88.00 & 12.00 & 87.00 & 99.00 & 87.20 & 6.00 & 93.30 & 72.00 & 52.00 \\
SqueezedAttn.
& 77.47 & 100.00 & 100.00 & 100.00 & 100.00 & 100.00 & 28.00 & 94.00 & 100.00 & 84.00 & 22.40 & 58.67 & 68.00 & 52.00 \\
FASA
& 77.44 & 100.00 & 100.00 & 96.00 & 100.00 & 96.00 & 40.00 & 90.00 & 100.00 & 84.80 & 17.20 & 66.67 & 60.00 & 56.00 \\
Loki
& 72.58 & 100.00 & 100.00 & 80.00 & 100.00 & 100.00 & 12.00 & 91.00 & 99.00 & 66.40 & 16.40 & 66.70 & 56.00 & 56.00 \\
\rowcolor{teal!10}
JSSA
& 82.55 & 100.00 & 100.00 & 100.00 & 100.00 & 100.00 & 88.00 & 94.00 & 100.00 & 83.20 & 24.00 & 64.00 & 68.00 & 52.00 \\
\midrule
\multicolumn{15}{c}{128K Context} \\
\midrule
FKV
& 76.97 & 100.00 & 100.00 & 100.00 & 100.00 & 84.00 & 64.00 & 85.00 & 98.00 & 89.60 & 0.00 & 84.00 & 60.00 & 36.00 \\
SnapKV
& 56.33 & 100.00 & 92.00 & 32.00 & 100.00 & 36.00 & 0.00 & 59.00 & 81.00 & 81.60 & 0.00 & 50.67 & 60.00 & 40.00 \\
QUEST
& 67.38 & 100.00 & 96.00 & 92.00 & 100.00 & 48.00 & 0.00 & 84.00 & 98.00 & 87.20 & 0.00 & 78.70 & 60.00 & 32.00 \\
SqueezedAttn.
& 68.96 & 100.00 & 100.00 & 96.00 & 100.00 & 76.00 & 0.00 & 87.00 & 97.00 & 79.20 & 0.00 & 57.33 & 64.00 & 40.00 \\
FASA
& 63.36 & 100.00 & 100.00 & 80.00 & 100.00 & 68.00 & 8.00 & 79.00 & 89.00 & 62.40 & 0.00 & 45.33 & 56.00 & 36.00 \\
Loki
& 37.98 & 88.00 & 76.00 & 0.00 & 84.00 & 24.00 & 0.00 & 36.00 & 43.00 & 38.40 & 0.40 & 40.00 & 48.00 & 16.00 \\
\rowcolor{teal!10}
JSSA
& 75.38 & 100.00 & 100.00 & 100.00 & 100.00 & 84.00 & 56.00 & 86.00 & 98.00 & 80.00 & 0.00 & 76.00 & 64.00 & 36.00 \\

\midrule
\multicolumn{15}{c}{\textit{Qwen3-14B}} \\
\midrule
\multicolumn{15}{c}{32K Context} \\
\midrule
FKV
& 91.77 & 100.00 & 100.00 & 100.00 & 100.00 & 100.00 & 96.00 & 92.00 & 100.00 & 100.00 & 86.40 & 94.67 & 64.00 & 60.00 \\
SnapKV
& 55.90 & 100.00 & 76.00 & 0.00 & 84.00 & 64.00 & 0.00 & 26.00 & 52.00 & 100.00 & 18.00 & 86.67 & 68.00 & 52.00 \\
QUEST
& 79.71 & 100.00 & 96.00 & 100.00 & 100.00 & 80.00 & 12.00 & 84.00 & 96.00 & 99.20 & 50.40 & 90.67 & 68.00 & 60.00 \\
SqueezedAttn.
& 88.22 & 100.00 & 100.00 & 100.00 & 100.00 & 96.00 & 68.00 & 93.00 & 99.00 & 100.00 & 78.80 & 92.00 & 64.00 & 56.00 \\
FASA
& 84.65 & 96.00 & 100.00 & 96.00 & 100.00 & 96.00 & 80.00 & 84.00 & 99.00 & 93.60 & 45.20 & 90.67 & 60.00 & 60.00 \\
Loki
& 90.18 & 100.00 & 100.00 & 100.00 & 100.00 & 100.00 & 92.00 & 99.00 & 96.00 & 100.00 & 68.00 & 89.33 & 72.00 & 56.00 \\
\rowcolor{teal!10}
JSSA
& 88.86 & 100.00 & 100.00 & 100.00 & 96.00 & 100.00 & 96.00 & 90.00 & 100.00 & 96.80 & 64.40 & 92.00 & 60.00 & 60.00 \\
\midrule
\multicolumn{15}{c}{64K Context} \\
\midrule
FKV
& 91.38 & 100.00 & 100.00 & 100.00 & 100.00 & 96.00 & 96.00 & 92.00 & 100.00 & 100.00 & 80.00 & 88.00 & 76.00 & 60.00 \\
SnapKV
& 60.49 & 100.00 & 88.00 & 4.00 & 100.00 & 56.00 & 4.00 & 29.00 & 69.00 & 100.00 & 24.40 & 76.00 & 76.00 & 60.00 \\
QUEST
& 80.58 & 100.00 & 100.00 & 100.00 & 100.00 & 68.00 & 36.00 & 92.00 & 90.00 & 100.00 & 49.60 & 76.00 & 80.00 & 56.00 \\
SqueezedAttn.
& 88.59 & 100.00 & 100.00 & 100.00 & 96.00 & 96.00 & 88.00 & 91.00 & 99.00 & 100.00 & 72.40 & 77.33 & 72.00 & 60.00 \\
FASA
& 84.08 & 100.00 & 100.00 & 96.00 & 100.00 & 96.00 & 88.00 & 83.00 & 96.00 & 98.40 & 43.60 & 76.00 & 60.00 & 56.00 \\
Loki
& 86.71 & 100.00 & 100.00 & 100.00 & 100.00 & 100.00 & 76.00 & 86.00 & 97.00 & 100.00 & 53.60 & 78.67 & 76.00 & 60.00 \\
\rowcolor{teal!10}
JSSA
& 88.47 & 100.00 & 100.00 & 100.00 & 100.00 & 96.00 & 96.00 & 93.00 & 98.00 & 100.00 & 60.40 & 74.67 & 72.00 & 60.00 \\
\midrule
\multicolumn{15}{c}{128K Context} \\
\midrule
FKV
& 83.75 & 100.00 & 100.00 & 100.00 & 100.00 & 92.00 & 72.00 & 78.00 & 98.00 & 99.20 & 49.60 & 80.00 & 64.00 & 56.00 \\
SnapKV
& 55.18 & 100.00 & 72.00 & 4.00 & 100.00 & 44.00 & 4.00 & 22.00 & 70.00 & 98.40 & 25.60 & 65.33 & 64.00 & 48.00 \\
QUEST
& 72.11 & 100.00 & 92.00 & 100.00 & 100.00 & 64.00 & 20.00 & 77.00 & 89.00 & 99.20 & 29.60 & 62.67 & 60.00 & 44.00 \\
SqueezedAttn.
& 78.75 & 100.00 & 96.00 & 100.00 & 100.00 & 92.00 & 52.00 & 75.00 & 97.00 & 99.20 & 51.20 & 57.33 & 52.00 & 52.00 \\
FASA
& 76.09 & 100.00 & 96.00 & 96.00 & 100.00 & 92.00 & 48.00 & 70.00 & 95.00 & 99.20 & 31.60 & 57.33 & 52.00 & 52.00 \\
Loki
& 76.43 & 100.00 & 100.00 & 96.00 & 100.00 & 100.00 & 32.00 & 81.00 & 90.00 & 99.20 & 42.00 & 65.33 & 44.00 & 44.00 \\
\rowcolor{teal!10}
JSSA
& 82.21 & 100.00 & 100.00 & 100.00 & 100.00 & 100.00 & 68.00 & 86.00 & 97.00 & 98.40 & 42.00 & 65.33 & 60.00 & 52.00 \\

\midrule
\multicolumn{15}{c}{\textit{Ministral-3-8B-Instruct-2512}} \\
\midrule
\multicolumn{15}{c}{32K Context} \\
\midrule
FKV
& 92.58 & 100.00 & 100.00 & 100.00 & 100.00 & 100.00 & 100.00 & 100.00 & 100.00 & 100.00 & 91.60 & 96.00 & 68.00 & 48.00 \\
SnapKV
& 65.75 & 100.00 & 100.00 & 44.00 & 96.00 & 44.00 & 0.00 & 43.00 & 98.00 & 97.60 & 34.80 & 81.33 & 68.00 & 48.00 \\
QUEST
& 86.51 & 100.00 & 100.00 & 100.00 & 100.00 & 100.00 & 56.00 & 98.00 & 100.00 & 100.00 & 64.00 & 94.67 & 68.00 & 44.00 \\
SqueezedAttn.
& 90.95 & 100.00 & 100.00 & 100.00 & 100.00 & 100.00 & 96.00 & 99.00 & 99.00 & 100.00 & 88.40 & 84.00 & 68.00 & 48.00 \\
FASA
& 78.11 & 96.00 & 80.00 & 84.00 & 96.00 & 76.00 & 36.00 & 93.00 & 87.00 & 99.20 & 77.60 & 86.67 & 60.00 & 44.00 \\
Loki
& 89.51 & 96.00 & 100.00 & 100.00 & 100.00 & 100.00 & 100.00 & 100.00 & 100.00 & 100.00 & 71.60 & 84.00 & 68.00 & 44.00 \\
\rowcolor{teal!10}
JSSA
& 90.72 & 96.00 & 100.00 & 100.00 & 100.00 & 100.00 & 100.00 & 100.00 & 100.00 & 100.00 & 86.00 & 81.33 & 68.00 & 48.00 \\
\midrule
\multicolumn{15}{c}{64K Context} \\
\midrule
FKV
& 89.45 & 88.00 & 96.00 & 100.00 & 100.00 & 100.00 & 100.00 & 100.00 & 100.00 & 100.00 & 78.80 & 88.00 & 68.00 & 44.00 \\
SnapKV
& 66.31 & 96.00 & 88.00 & 48.00 & 92.00 & 72.00 & 0.00 & 63.00 & 97.00 & 100.00 & 30.00 & 72.00 & 60.00 & 44.00 \\
QUEST
& 82.95 & 92.00 & 96.00 & 100.00 & 100.00 & 88.00 & 60.00 & 98.00 & 100.00 & 100.00 & 52.40 & 76.00 & 72.00 & 44.00 \\
SqueezedAttn.
& 87.60 & 80.00 & 88.00 & 100.00 & 100.00 & 96.00 & 96.00 & 98.00 & 100.00 & 100.00 & 88.80 & 76.00 & 68.00 & 48.00 \\
FASA
& 80.45 & 72.00 & 92.00 & 100.00 & 100.00 & 80.00 & 44.00 & 95.00 & 98.00 & 100.00 & 80.80 & 72.00 & 64.00 & 48.00 \\
Loki
& 87.82 & 96.00 & 96.00 & 100.00 & 100.00 & 92.00 & 96.00 & 99.00 & 100.00 & 100.00 & 72.00 & 82.67 & 64.00 & 44.00 \\
\rowcolor{teal!10}
JSSA
& 89.02 & 92.00 & 96.00 & 100.00 & 100.00 & 100.00 & 100.00 & 98.00 & 100.00 & 100.00 & 83.20 & 76.00 & 68.00 & 44.00 \\
\midrule
\multicolumn{15}{c}{128K Context} \\
\midrule
FKV
& 85.93 & 96.00 & 100.00 & 100.00 & 96.00 & 100.00 & 100.00 & 100.00 & 94.00 & 100.00 & 40.40 & 94.67 & 60.00 & 36.00 \\
SnapKV
& 62.84 & 92.00 & 96.00 & 40.00 & 84.00 & 56.00 & 0.00 & 71.00 & 92.00 & 100.00 & 23.20 & 78.67 & 52.00 & 32.00 \\
QUEST
& 79.02 & 96.00 & 100.00 & 100.00 & 96.00 & 96.00 & 36.00 & 100.00 & 97.00 & 100.00 & 35.60 & 86.67 & 48.00 & 36.00 \\
SqueezedAttn.
& 82.85 & 88.00 & 100.00 & 100.00 & 96.00 & 92.00 & 80.00 & 99.00 & 98.00 & 100.00 & 60.00 & 72.00 & 60.00 & 32.00 \\
FASA
& 72.81 & 96.00 & 88.00 & 88.00 & 80.00 & 56.00 & 48.00 & 83.00 & 84.00 & 100.00 & 60.80 & 66.67 & 56.00 & 40.00 \\
Loki
& 81.84 & 96.00 & 96.00 & 100.00 & 92.00 & 88.00 & 76.00 & 95.00 & 95.00 & 100.00 & 51.20 & 82.67 & 60.00 & 32.00 \\
\rowcolor{teal!10}
JSSA
& 85.97 & 92.00 & 100.00 & 100.00 & 96.00 & 100.00 & 100.00 & 99.00 & 97.00 & 100.00 & 61.60 & 76.00 & 60.00 & 36.00 \\
\bottomrule
\end{tabular}
}
\end{table}

\newpage

\subsection{Complete Reasoning Results}
\label{app:reasoning_results}

Tables~\ref{tab:reasoning_qwen_accuracy}--\ref{tab:reasoning_glm_accuracy} report pass@1 accuracy across KV budgets of 512, 1,024, and 2,048 tokens.
Table~\ref{tab:reasoning_length_full} separately reports the corresponding generation lengths relative to full-KV attention.
Full-KV attention results are independent of the sparse token budget and are therefore reported once.
Across models and benchmarks, JSSA remains comparatively stable across budgets, whereas competing methods often require larger budgets to recover accuracy and reduce generation-length inflation.

\begin{table}[h]
\centering
\caption{Complete reasoning accuracy on Qwen3-8B across KV budgets. FKV is reported once because it does not depend on the sparse token budget.}
\label{tab:reasoning_qwen_accuracy}
\footnotesize
\setlength{\tabcolsep}{4pt}
\begin{tabular}{lccc|ccc|ccc}
\toprule
& \multicolumn{3}{c}{AIME 2025}
& \multicolumn{3}{c}{MATH-500}
& \multicolumn{3}{c}{GPQA-Diamond} \\
\cmidrule(lr){2-4}
\cmidrule(lr){5-7}
\cmidrule(lr){8-10}
Method
& 512 & 1024 & 2048
& 512 & 1024 & 2048
& 512 & 1024 & 2048 \\
\midrule
FKV
& \multicolumn{3}{c}{69.58}
& \multicolumn{3}{c}{94.53}
& \multicolumn{3}{c}{57.95} \\
TriAttention
& 5.83 & 20.00 & 37.50
& 73.80 & 86.20 & 93.20
& 31.82 & 40.40 & 50.13 \\
QUEST
& 47.08 & 62.92 & 66.25
& 92.73 & 93.50 & 89.93
& 57.01 & 58.14 & 56.12 \\
SeerAttention-R
& 26.67 & 42.08 & 55.00
& 85.70 & 91.80 & 94.03
& 54.67 & 55.93 & 58.52 \\
FASA
& 30.83 & 60.42 & 67.50
& 87.93 & 93.58 & 94.60
& 56.06 & 57.45 & 59.97 \\
\rowcolor{teal!10}
JSSA
& 64.58 & 68.33 & 67.92
& 94.05 & 94.23 & 94.45
& 58.08 & 57.39 & 58.21 \\
\bottomrule
\end{tabular}
\end{table}

\begin{table}[h]
\centering
\caption{
Complete reasoning accuracy on Gemma4-26B-A4B-It across KV budgets.
FKV is reported once because it is independent of the sparse token budget.
}
\label{tab:reasoning_gemma_accuracy}
\footnotesize
\setlength{\tabcolsep}{4pt}
\begin{tabular}{lccc|ccc|ccc}
\toprule
& \multicolumn{3}{c}{AIME 2025}
& \multicolumn{3}{c}{MATH-500}
& \multicolumn{3}{c}{GPQA-Diamond} \\
\cmidrule(lr){2-4}
\cmidrule(lr){5-7}
\cmidrule(lr){8-10}
Method
& 512 & 1024 & 2048
& 512 & 1024 & 2048
& 512 & 1024 & 2048 \\
\midrule
FKV
& \multicolumn{3}{c}{90.00}
& \multicolumn{3}{c}{97.15}
& \multicolumn{3}{c}{80.87} \\
TriAttention
& 13.33 & 22.50 & 54.17
& 68.35 & 83.95 & 95.35
& 40.28 & 43.94 & 68.69 \\
QUEST
& 28.75 & 50.42 & 70.42
& 88.20 & 93.65 & 96.33
& 63.07 & 71.40 & 77.84 \\
FASA
& 57.92 & 73.75 & 84.17
& 95.25 & 97.08 & 97.13
& 78.91 & 80.56 & 80.87 \\
\rowcolor{teal!10}
JSSA
& 88.75 & 86.67 & 88.33
& 96.18 & 97.08 & 97.10
& 80.68 & 81.94 & 81.12 \\
\bottomrule
\end{tabular}
\end{table}

\begin{table}[h]
\centering
\caption{
Complete reasoning accuracy on GLM-4.7-Flash across KV budgets.
FKV is reported once because it is independent of the sparse token budget.
}
\label{tab:reasoning_glm_accuracy}
\footnotesize
\setlength{\tabcolsep}{4pt}
\begin{tabular}{lccc|ccc|ccc}
\toprule
& \multicolumn{3}{c}{AIME 2025}
& \multicolumn{3}{c}{MATH-500}
& \multicolumn{3}{c}{GPQA-Diamond} \\
\cmidrule(lr){2-4}
\cmidrule(lr){5-7}
\cmidrule(lr){8-10}
Method
& 512 & 1024 & 2048
& 512 & 1024 & 2048
& 512 & 1024 & 2048 \\
\midrule
FKV
& \multicolumn{3}{c}{90.42}
& \multicolumn{3}{c}{96.80}
& \multicolumn{3}{c}{75.44} \\
FASA
& 37.08 & 65.00 & 81.25
& 91.08 & 95.65 & 96.80
& 55.30 & 69.38 & 73.61 \\
\rowcolor{teal!10}
JSSA
& 86.67 & 91.67 & 91.25
& 96.95 & 96.88 & 96.88
& 75.25 & 77.53 & 78.03 \\
\bottomrule
\end{tabular}
\end{table}

\begin{table}[h]
\centering
\caption{Generation length relative to FKV across reasoning benchmarks and KV budgets. Lower is better.}
\label{tab:reasoning_length_full}
\footnotesize
\setlength{\tabcolsep}{5pt}
\begin{tabular}{lllccc}
\toprule
Model & Benchmark & Method & 512 & 1024 & 2048 \\
\midrule

\multirow{15}{*}{Qwen3-8B}
& \multirow{5}{*}{AIME 2025}
& TriAttention & 2.07$\times$ & 1.75$\times$ & 1.54$\times$ \\
& & QUEST & 1.51$\times$ & 1.29$\times$ & 1.16$\times$ \\
& & SeerAttention-R & 1.76$\times$ & 1.54$\times$ & 1.30$\times$ \\
& & FASA & 1.71$\times$ & 1.30$\times$ & 1.10$\times$ \\
& & \cellcolor{teal!10}JSSA
& \cellcolor{teal!10}1.16$\times$
& \cellcolor{teal!10}1.07$\times$
& \cellcolor{teal!10}1.04$\times$ \\
\cmidrule(lr){2-6}
& \multirow{5}{*}{MATH-500}
& TriAttention & 2.59$\times$ & 1.70$\times$ & 1.22$\times$ \\
& & QUEST & 1.38$\times$ & 1.21$\times$ & 1.27$\times$ \\
& & SeerAttention-R & 1.91$\times$ & 1.35$\times$ & 1.14$\times$ \\
& & FASA & 1.80$\times$ & 1.17$\times$ & 1.07$\times$ \\
& & \cellcolor{teal!10}JSSA
& \cellcolor{teal!10}1.23$\times$
& \cellcolor{teal!10}1.09$\times$
& \cellcolor{teal!10}1.07$\times$ \\
\cmidrule(lr){2-6}
& \multirow{5}{*}{GPQA-Diamond}
& TriAttention & 4.82$\times$ & 4.13$\times$ & 3.33$\times$ \\
& & QUEST & 1.20$\times$ & 1.03$\times$ & 0.93$\times$ \\
& & SeerAttention-R & 2.01$\times$ & 1.43$\times$ & 1.09$\times$ \\
& & FASA & 1.18$\times$ & 1.06$\times$ & 1.03$\times$ \\
& & \cellcolor{teal!10}JSSA
& \cellcolor{teal!10}1.06$\times$
& \cellcolor{teal!10}1.02$\times$
& \cellcolor{teal!10}1.01$\times$ \\

\midrule

\multirow{12}{*}{Gemma4}
& \multirow{4}{*}{AIME 2025}
& TriAttention & 2.42$\times$ & 2.29$\times$ & 1.75$\times$ \\
& & QUEST & 2.18$\times$ & 1.96$\times$ & 1.57$\times$ \\
& & FASA & 1.86$\times$ & 1.38$\times$ & 1.17$\times$ \\
& & \cellcolor{teal!10}JSSA
& \cellcolor{teal!10}1.21$\times$
& \cellcolor{teal!10}1.17$\times$
& \cellcolor{teal!10}1.09$\times$ \\
\cmidrule(lr){2-6}
& \multirow{4}{*}{MATH-500}
& TriAttention & 3.50$\times$ & 2.37$\times$ & 1.37$\times$ \\
& & QUEST & 2.08$\times$ & 1.65$\times$ & 1.28$\times$ \\
& & FASA & 1.52$\times$ & 1.14$\times$ & 1.08$\times$ \\
& & \cellcolor{teal!10}JSSA
& \cellcolor{teal!10}1.33$\times$
& \cellcolor{teal!10}1.13$\times$
& \cellcolor{teal!10}1.08$\times$ \\
\cmidrule(lr){2-6}
& \multirow{4}{*}{GPQA-Diamond}
& TriAttention & 3.32$\times$ & 3.06$\times$ & 2.17$\times$ \\
& & QUEST & 2.33$\times$ & 2.02$\times$ & 1.53$\times$ \\
& & FASA & 1.44$\times$ & 1.18$\times$ & 1.06$\times$ \\
& & \cellcolor{teal!10}JSSA
& \cellcolor{teal!10}1.25$\times$
& \cellcolor{teal!10}1.12$\times$
& \cellcolor{teal!10}1.07$\times$ \\

\midrule

\multirow{6}{*}{GLM-4.7-Flash}
& \multirow{2}{*}{AIME 2025}
& FASA & 3.30$\times$ & 2.24$\times$ & 1.58$\times$ \\
& & \cellcolor{teal!10}JSSA
& \cellcolor{teal!10}1.31$\times$
& \cellcolor{teal!10}1.17$\times$
& \cellcolor{teal!10}1.10$\times$ \\
\cmidrule(lr){2-6}
& \multirow{2}{*}{MATH-500}
& FASA & 3.79$\times$ & 1.90$\times$ & 1.28$\times$ \\
& & \cellcolor{teal!10}JSSA
& \cellcolor{teal!10}1.18$\times$
& \cellcolor{teal!10}1.10$\times$
& \cellcolor{teal!10}1.05$\times$ \\
\cmidrule(lr){2-6}
& \multirow{2}{*}{GPQA-Diamond}
& FASA & 2.75$\times$ & 2.30$\times$ & 1.75$\times$ \\
& & \cellcolor{teal!10}JSSA
& \cellcolor{teal!10}1.60$\times$
& \cellcolor{teal!10}1.38$\times$
& \cellcolor{teal!10}1.23$\times$ \\

\bottomrule
\end{tabular}
\end{table}

JSSA is substantially less sensitive to the token budget than competing methods.
Even at 512 tokens, it remains close to full-KV accuracy on most model--benchmark pairs, while several baselines exhibit severe accuracy loss and generation-length inflation.
Increasing the budget generally narrows these gaps, but JSSA continues to provide the most consistent trade-off between final-answer accuracy and reasoning length.

\end{document}